%% file: main.tex
\documentclass[10pt,twocolumn,letterpaper]{article}

\usepackage{iccv}              % To produce the CAMERA-READY version

\definecolor{iccvblue}{rgb}{0.21,0.49,0.74}
\usepackage[pagebackref,breaklinks,colorlinks,allcolors=iccvblue]{hyperref}
\usepackage{fontawesome}
\usepackage{pifont} % 加入到 LaTeX 导言区
\newcommand{\cmark}{\ding{51}} % ✔️
\newcommand{\xmark}{\ding{55}} % ❌
\usepackage{multirow}

\def\paperID{4216} % *** Enter the Paper ID here
\def\confName{ICCV}
\def\confYear{2025}

\title{\textit{Where, What, Why:} Towards Explainable Driver Attention Prediction} %via Large Multi-modal Model}
\author{Yuchen Zhou\textsuperscript{1,2}\thanks{Equal Contribution.}, Jiayu Tang\textsuperscript{1}\footnotemark[1], Xiaoyan Xiao\textsuperscript{1}, Yueyao Lin\textsuperscript{1}, 
Linkai Liu\textsuperscript{1}, Zipeng Guo\textsuperscript{1}, \\ Hao Fei\textsuperscript{2}, Xiaobo Xia\textsuperscript{2}, Chao Gou\textsuperscript{1}\thanks{Corresponding Author.}\\
\textsuperscript{1}Sun Yat-sen University, \textsuperscript{2}National University of Singapore\\
{\tt\small https://github.com/yuchen2199/Explainable-Driver-Attention-Prediction}
}

\begin{document}
\maketitle

%\thanks{$\dagger$ Equal contribution. * Corresponding author.}

% \footnotetext{
% \\
% \textsuperscript{*} Equal Contribution \\
% \textsuperscript{†} Corresponding Author
% }

\input{sec/0_abstract}    
\input{sec/1_intro}
\input{sec/2_relatedwork}

\input{sec/3_dataset}

\input{sec/4_method}

\input{sec/5_experiment}

\input{sec/6_conclusion}

{
    \small
\bibliographystyle{ieeenat_fullname}
    \bibliography{main}
}
    
\input{sec/7S_dataset}
\input{sec/8S_model}

\end{document}

%% file: sec/0_abstract.tex
\begin{abstract}

Modeling task-driven attention in driving is a fundamental challenge for both autonomous vehicles and cognitive science. Existing methods primarily predict where drivers look by generating spatial heatmaps, but fail to capture the cognitive motivations behind attention allocation in specific contexts, which limits deeper understanding of attention mechanisms. To bridge this gap, we introduce Explainable Driver Attention Prediction, a novel task paradigm that jointly predicts spatial attention regions (\textbf{where}), parses attended semantics (\textbf{what}), and provides cognitive reasoning for attention allocation (\textbf{why}). 
To support this, we present \textbf{W³DA}, the first large-scale explainable driver attention dataset. It enriches existing benchmarks with detailed semantic and causal annotations across diverse driving scenarios, including normal conditions, safety-critical situations, and traffic accidents. We further propose \textbf{LLada}, a \textbf{L}arge \textbf{La}nguage model-driven framework for \textbf{d}river \textbf{a}ttention prediction, which unifies pixel modeling, semantic parsing, and cognitive reasoning within an end-to-end architecture. Extensive experiments demonstrate the effectiveness of LLada, exhibiting robust generalization across datasets and driving conditions. 
This work serves as a key step toward a deeper understanding of driver attention mechanisms, with significant implications for autonomous driving, intelligent driver training, and human-computer interaction. 

\end{abstract}

%% file: sec/1_intro.tex
\section{Introduction}
\label{sec:intro}

% TODO 展现一个特别好的例子，来展现我们的where-what-why的make sense。
% 找一个同时体现 交通驾驶领域知识+context场景上下文的

% As an old English proverb goes，"眼睛是心灵的窗户。" 建模并理解人的视觉注意可以揭示人的认知能力、经验知识和决策策略，特别是在人类执行特定的复杂任务时，例如驾驶。这对于发展psychology and cognitive science、增强人机交互、构建类人的智能体具有巨大的潜力。在基于深度学习的显著性检测范式驱动下，driver attention prediction任务被提出，并取得巨大进展。然而现有的范式局限于“modelling”，也就是只能回答预测人的注意力“在哪里”。而不能做到真正的“understanding”，例如将驾驶注意从隐式的眼动行为转换为显式的明确的解释，也就是不能回答“what”和“why”。这种显式的明确的解释的缺失，同时阻碍了更深入的认知理论研究和实际落地应用。

As an old proverb goes, \textit{``The eyes are the window to the soul.''} Human eyes serve as a crucial gateway to cognition, reflecting how human perceive, interpret, and interact with their environment \cite{posner1990attention, zhang2020human,mele2012gaze,zhou2023learning, stephenson2021gaze,ijcai2024p882,zhou2024learning,song2024vitgaze}. 
Modeling and understanding human visual attention provide fundamental insights into cognitive abilities, experiential knowledge, and decision-making strategies, especially in complex, task-driven scenarios such as driving \cite{huang2018predicting, chen2020air, liu2021goal, kotseruba2022practical, pettine2023human,2024_IV_data,zhou2024hktsg,huang2024driver,zhou2023pit}. 

% This holds great promise for advancing cognitive science \cite{mele2012gaze,stephenson2021gaze}, enhancing human-computer interaction \cite{wang2024g,kim2024enhancing,huang2018predicting}, and enabling the development of human-like intelligent agents \cite{zhao2022human,zhao2023improving, yan2025voila}.

% Recently, driver attention prediction has emerged as a key research direction and has achieved remarkable progress. 然而，现有的研究范式仅仅聚焦于回答驾驶员们会看哪里（粗浅的隐式的建模attention机制），但却没有真正的理解驾驶员在特定场景上下文下所产生注意力的更深层动机，也就是显式的回答注意力区域是什么和为什么会产生这样的注意力。这种对于深层动机的理解的缺失，同时阻碍了更深入的认知理论研究和实际落地应用。

%先前的研究仅仅能回答驾驶员们会关注哪里（粗浅的隐式的建模attention机制），而我们工作能同时回答驾驶员会关注哪里，在关注什么以及为什么关注（以一种更加深入的更加显式的方式来建模驾驶attention机制，挖掘注意力分配背后的所反应的交通规则知识、驾驶安全感知和驾驶目标驱动等）。我们工作向社区提供揭开注意力机制的秘密的新窗口。

\begin{figure}[t]
	\includegraphics[trim=15 20 15 15, clip=true, width=0.46\textwidth]{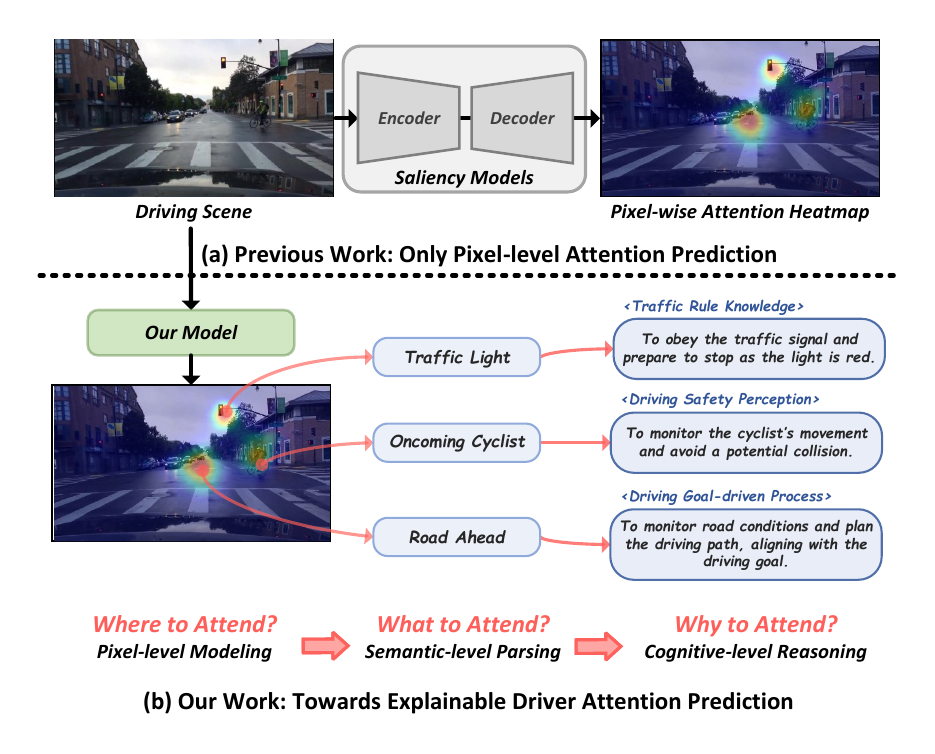}
	\caption{Prior studies focus only on predicting where drivers allocate their attention, offering an implicit and shallow representation of attention. In contrast, our work proposes a more explicit and in-depth paradigm that answers where to attend, what to attend, and why attention is allocated there, revealing underlying factors behind attention allocation such as traffic rule knowledge, driving safety perception, and goal-driven behaviors. Our work unlocks a deeper understanding of driver attention mechanisms, providing new insights to the research community.}
	\label{fig1}
\end{figure}

In recent years, driver attention prediction has emerged as a key research direction and has achieved remarkable progress \cite{Palazzi2019dreyeve, xia2019predicting, dua2020dgaze, pal2020looking,gou2022cascade,huang2024task}. 
Existing paradigms \cite{LBW, fang2022DADA, chen2023fblnet, zhu2023attniccv,zhao2025salm2,zhou2025behavior} primarily focus on predicting \textit{where} drivers look by generating spatial heatmaps, as shown in Fig. \ref{fig1}(a). These methods fundamentally perform pixel-space regression, offering only a shallow and implicit representation of attention mechanisms.
However, these approaches fail to capture the underlying cognitive motivations behind attention allocation in specific driving contexts. For instance, in the scenario depicted in Fig. \ref{fig1}, a driver attends to the red traffic light to comply with traffic rules, monitors an oncoming cyclist at an intersection to ensure driving safety, and observes the road ahead to navigate toward the intended destination. 
Unfortunately, no existing work has effectively modeled these semantic-level (i.e., \textit{what} is being attended to) and cognitive-level (i.e., \textit{why} attention is allocated) aspects of driver attention. This critical limitation hinders both a deeper understanding of driver attention mechanisms and the real-world applicability of these models.

% To bridge these critical gaps, 我们提出了一个统一架构来同时实现对驾驶注意的像素级区域（where）、语义级解释（what）、认知推理成因（why）的预测，以穿越像素空间、语义空间、认知知识空间实现对驾驶员注意力的更全面的可解释的建模和理解。具体的，我们的工作同现有的注意力预测工作有几大关键提升： (1)我们提出了一个大规模的可解释驾驶注意数据集，W3DA，它是semantic and reasoning-aware的，在跨越现有多个驾驶注意数据集上注释了注意区域的语义和成因。这些注释可以作为今后的全面的可解释的注意力理解工作的强有力的基础，以促进认知科学研究和实际落地（例如自动驾驶的可解释决策提高用户可信，以及智能高效的驾驶教学等等）。（2）我们提出了第一个通用的模型架构以实现对驾驶注意的区域（where）、语义（what）、成因（why）的同时预测。(3)虽然现有的模型针对的是单个特定数据集进行训练和测试，但我们通过我们提出的整合现有主流数据集的W3DA进行跨数据集协同训练，实现了对正常驾驶、突发场景和交通事故等场景的统一建模，improving model generalization and real-world applicability. 

To bridge these critical gaps, we propose a novel unified task paradigm, termed \textbf{Explainable Driver Attention Prediction}, as shown in Fig. \ref{fig1}(b). This paradigm requires models to simultaneously predict pixel-level attention regions (answering \textbf{\textit{where}}), provide semantic-level parsing (answering \textbf{\textit{what}}), and perform cognitive-level reasoning to explain the underlying causes of attention allocation (answering \textbf{\textit{why}}). By integrating spatial, semantic, and cognitive knowledge, this enables a more comprehensive and explainable modeling of driver attention.

Specifically, our work introduces several key advancements over existing driver attention prediction methods:  
\textbf{(1)} We introduce \textbf{W³DA}, a large-scale explainable driver attention dataset, which integrates multiple mainstream driver attention benchmarks and incorporates additional annotations for semantic parsing (\textit{what}) and cognitive reasoning (\textit{why}). It provides spatial, semantic, and cognitive labels for attention regions, explicitly enabling \textit{where, what, and why} reasoning in driver attention modeling.
\textbf{(2)} We propose \textbf{LLada}, a \textbf{L}arge \textbf{la}nguage model-driven framework for \textbf{d}river \textbf{a}ttention prediction, which is the first unified architecture capable of jointly predicting attention regions (\textit{where}), semantic interpretations (\textit{what}), and cognitive reasoning (\textit{why}) in driver attention modeling.  
\textbf{(3)} Unlike existing models trained and evaluated separately on individual datasets, we leverage the W³DA dataset for cross-domain collaborative training, enabling a unified modeling approach for normal driving, safety-critical situations, and traffic accidents. % This approach significantly enhances model generalization and real-world applicability.

% 实验结果+拔立意（认知科学+实际应用？）
% 我们实施了全面的实验来验证我们框架的有效性，实验证明我们的框架在W3DA上超越了所有sota方法，跨越normal driving, critical situations, and traffic accidents等多个场景。不仅如此，仅在W3DA上训练的LLADA还超越了在DR(eye)VE、LBW、BDDA、DADA-2000等数据集上超越了先前所有的单一专有模型，体现出我们框架强大的泛化性和通用性。同时，我们还进行了大量的人类评价，定性证明了我们框架的优越性。我们的框架跨越serve as a strong foundation for future research on explainable attention modeling, facilitating advancements in cognitive science (研究任务驱动下的注意力分配的深层机理) and real-world applications such as interpretable decision-making in autonomous driving and intelligent driver training systems.

% 大量的实验验证了我们提出的Llada的有效性，无论是注意力图预测指标还是解释文本生成指标，我们的框架都在W3DA上超越了所有sota方法，跨越normal driving, safety critical situations, and traffic accidents等多类场景。不仅如此，仅在W3DA上训练的LLADA在DR(eye)VE、LBW、BDDA、DADA-2000等数据集上超越了大部分先前单一训练的专有模型，实现极具竞争力的性能表现，体现出我们框架强大的泛化性和通用性。此外，我们还进行了大量的定性分析和人类评价，充分证明我们llada模型对人类驾驶员注意机制的准确建模及高度对齐。我们的框架将serve as a strong foundation for future research on comprehensive and explainable attention modeling, facilitating advancements in cognitive science and real-world applications such as interpretable decision-making in autonomous driving and intelligent driver training systems.

Extensive experiments validate the effectiveness of LLada, demonstrating its superior performance over all state-of-the-art (SOTA) methods on W³DA in both attention map prediction and textual explanation generation. It consistently excels across various driving scenarios.
Moreover, trained solely on W³DA, LLada outperforms most independently trained specialized models across the entire DR(eye)VE \cite{Palazzi2019dreyeve}, %LBW \cite{LBW}, 
BDDA \cite{xia2019predicting}, and DADA-2000 \cite{fang2019dada} datasets, achieving highly competitive performance and demonstrating strong generalization.
Further qualitative analyses % and human evaluations 人类评估还没完成 
confirm its strong alignment with human driver attention. Our work establishes a solid foundation for comprehensive and explainable attention modeling, paving the way for advancements in cognitive science and real-world applications such as interpretable autonomous driving and intelligent driver training.
Before delving into details, we clearly emphasize our contributions as follows:
\begin{itemize}
    \item \textbf{(Paradigm)} We introduce Explainable Driver Attention Prediction, a novel paradigm that extends traditional driver attention prediction by jointly reasoning over \textit{where}, \textit{what}, and \textit{why} to achieve a more comprehensive understanding of attention in driving.
    \item \textbf{(Dataset)} We present W³DA, a large-scale explainable driver attention dataset, which enhances existing benchmarks with semantic (\textit{what}) and causal (\textit{why}) annotations, spanning normal, safety-critical, and accident scenarios. 
    \item \textbf{(Methodology)} We propose LLada, the first large language model-driven driver attention prediction framework, enabling end-to-end training and joint prediction of attention regions (\textit{where}), semantic parsing (\textit{what}), and cognitive reasoning (\textit{why}).
    \item \textbf{(Experiments)} Extensive experiments demonstrate that LLada consistently outperforms existing SOTA methods across multiple tasks, scenarios, metrics, and settings, demonstrating superior robustness and generalization.
\end{itemize}

%-------------------------------------------------------------------------

% \begin{figure*}
%   \centering
%   \begin{subfigure}{0.68\linewidth}
%     \fbox{\rule{0pt}{2in} \rule{.9\linewidth}{0pt}}
%     \caption{An example of a subfigure.}
%     \label{fig:short-a}
%   \end{subfigure}
%   \hfill
%   \begin{subfigure}{0.28\linewidth}
%     \fbox{\rule{0pt}{2in} \rule{.9\linewidth}{0pt}}
%     \caption{Another example of a subfigure.}
%     \label{fig:short-b}
%   \end{subfigure}
%   \caption{Example of a short caption, which should be centered.}
%   \label{fig:short}
% \end{figure*}

%% file: sec/2_relatedwork.tex
\section{Related Work}
\label{sec:RelatedWork}

\subsection{Driver Attention Prediction}

% 要写出好的Related work，步骤为：(1) 首先，列出和自己论文的方法比较相关的论文。（**Related work中最重要的部分**，如果没讨论，一些reviewer会直接以此拒掉论文）(2) 然后，根据论文的研究方向和算法技术来确定Related work要讨论哪几个topics，列出几个topics下面要讨论的论文(3) 最后，基于前两步列出的论文组织related work的写作思路。

% 先总说一下注意力 highlevel的动机

% 注意力数据集
\noindent \textbf{Datasets.}
% 现状+highlight我们的独特贡献
% DREYEVE LBW lvGAZE -> BDDA -> DADA 
% 专门数据集显著地支持了驾驶员注意力研究的发展。最初的贡献，如DR（eye）VE，引入了在一个通畅的交通环境中收集和分析司机注视的综合框架。他们在一辆汽车上安装了眼球追踪装置，记录下了八名司机的注意力。类似的，LBW提供了28名司机在真实驾驶过程中的人脸、场景图像与凝视信息。Cheng等人介绍了IVGaze [4]，一个在真实驾驶条件下的综合车内注视数据集，但并不包含对应的driving footage。另外，有一些特殊场景的驾驶注意力在实验室环境中采集，例如Xia等人[20]开发了BDD-A，重点关注制动事件和拥挤的交通状况的视频剪辑。DADA-2000已经扩大了重点到包括交通事故分析，有54种事故的2000个视频剪辑和相应的实验室内注意力数据，为探索复杂的驾驶员注意动态提供了更丰富的数据。
% 尽管大量的驾驶注意数据集被提出，包含driving footage and gaze information，但依旧没有一个数据集来探究注意力分配背后的semantic interpretation and causal reasoning，限制了对驾驶注意力机制的深入理解。因此，在本文中，我们提出W3DA, a large-scale explainable driver attention dataset that provides fine-grained labels for attention regions along with their semantic meanings and underlying causes, making it the first dataset to explicitly support where, what, and why reasoning in driver attention modeling.
Specialized driver attention datasets have significantly advanced the field. Early works like DR(eye)VE \cite{Palazzi2019dreyeve} introduced a pipeline to analyze driver gaze in normal traffic, recording attention data from eight drivers. Similarly, the LBW dataset \cite{LBW} provided gaze data collected from 28 drivers in real driving conditions. Cheng et al. introduced IVGaze \cite{cheng2024you}, which collected in-vehicle gaze data but lacked corresponding driving footage. The BDD-A dataset by Xia et al. \cite{xia2019predicting} focused on safety-critical events like emergency braking and traffic congestion. DADA-2000 \cite{fang2019dada} extended this to traffic accidents, including 2,000 videos and corresponding gaze data.
While existing datasets have provided large-scale driver attention data, they primarily associate spatial heatmaps with driving scenes without capturing the cognitive reasoning behind attention allocation, which limits a deeper understanding of driver attention mechanisms. To address this, we introduce W³DA, the first large-scale dataset for explainable driver attention modeling. W³DA extends beyond heatmaps by providing fine-grained annotations that link attention heatmaps to their semantic meaning and underlying causes, offering unprecedented insights into driver attention mechanisms.

% 注意力方法
\noindent \textbf{Models.}
% 现状+highlight我们的独特贡献
%在早期，主要的方法主要采用概率模型来预测注意力。
Attention prediction models have evolved from early image processing techniques like ITTI \cite{itti1998model} and GBVS \cite{harel2006graph} to deep learning-based architectures. CNNs became fundamental, enabling richer visual feature representations. Palazzi et al. \cite{Palazzi2019dreyeve} integrated RGB, optical flow, and segmentation, while Xia et al. \cite{xia2019predicting} employed convolutional LSTMs. Deng et al. \cite{deng2020CDNN} proposed a fully convolutional encoder-decoder. Multi-resolution networks \cite{hu2021data} and heterogeneous networks \cite{hu2022novel} further incorporated multilevel visual content. ASIAFNet \cite{li2022adaptive} emphasized semantic-based object-level attention.
Beyond CNNs,  Inverse Reinforcement Learning \cite{baee2021medirl} and Generative Adversarial Networks \cite{araluce2022aragan, baee2021medirl} have been explored to model driver attention. Amadori et al. \cite{amadori2021hammerdrive} incorporated driving inputs and vehicle states in VR simulations. Unsupervised methods \cite{zhu2023attniccv} explored scene dynamics without reliance on labeled attention data. Shi et al. \cite{shi2023fixated, shi2024weakly} proposed FOD-Net with saliency guidance. Recent advances integrate Transformers \cite{VIT} to enhance global context modeling. ACT-Net \cite{gou2022driver} pioneered CNN-Transformer fusion, followed by FBLNet \cite{chen2023fblnet}, which introduces a feedback loop mechanism for experience accumulation, further bridging attention modeling with human-like learning.
Despite these advancements, existing models still fail to explicitly capture the cognitive motivations behind driver attention. To address this limitation, we propose LLada, a unified end-to-end framework that jointly models spatial (\textit{where}), semantic (\textit{what}), and cognitive (\textit{why}) factors, enabling a more comprehensive and explainable approach to driver attention prediction.

% 需要把测评中用到的所有方法全部cover到

\subsection{Multimodal Large Language Models}
% mllm取得卓越进展 llava等
% -> 自动驾驶上的mllm （简要）
% lisa等工作的思想 （review）
% -> 眼动上的mllm （nips2024，gazexplain）
% 值得一提的是gazexplain，然而使用的是blip，解释也是聚焦于what而不是更深层的why
%Voila-A: Aligning Vision-Language Models with User's Gaze Attention
% highlight 独特贡献

%近年，以大规模语言模型（LLM）为基础的多模态大模型在许多视觉任务中取得了显著进展，跨越image-level captioning、object-level detection和pixel-level segmentation。特别是，LLM的强大推理能力被成功引入视觉任务中，推动了视觉认知能力的飞跃。这些模型在复杂视觉认知任务中展现出了优越的表现，尤其是在理解图像内容、生成文本描述以及多模态任务的融合上。遗憾的是，在当前在注意力/显著性预测领域，据我们所知尚未有研究探讨如何基于LLM构建具有推理能力的注意力/显著性计算模型，严重限制了对注意力机制的背后成因的深层建模和理解。

% 我们的工作受到MLLM最新进展的启发，提出（）

% Recent advancements in multimodal large language models (MLLMs) have demonstrated remarkable progress across various-level tasks, spanning image-level captioning, object-level detection, and pixel-level segmentation, 同时还在自动驾驶、机器人等实际应用上大放异彩. Notably, the powerful reasoning capabilities of large language models (LLMs) have been successfully introduced into vision tasks, leading to a paradigm shift in visual cognition. 
% 一些工作开始尝试将人类注意信息并入到MLLM中，Voila-A利用从 AR/VR 环境中收集的注视信息输入到MLLM中，在需要用户特定视觉焦点的任务中表现出色

Recent advancements in multimodal large language models (MLLMs) \cite{liu2023visual,zhang2024llava,liu2024improved,yang2024qwen2,wu2024next,lmeye,li2025perception,yang2023continual} have driven significant progress across image-level captioning \cite{chen2025sharegpt4video}, object-level detection \cite{zang2024contextual,bianchi2024devil}, and pixel-level segmentation \cite{lai2024lisa,xia2024gsva,ren2024pixellm}, with notable applications in autonomous driving \cite{cui2024survey, tian2024drivevlm, sima2024drivelm,zhang2023bev} and robotics \cite{ long2024robollm,lin2024airvista,yuan2025seeing,tian2025uavs}. 
The powerful reasoning capabilities of large language models (LLMs) have been successfully introduced into vision tasks, fostering richer multimodal interactions and enhancing visual reasoning capabilities.
Building on this foundation, recent studies have explored incorporating human attention as an additional modality to enhance model performance and interpretability. Voila-A \cite{yan2025voila} aligned MLLMs with the human gaze (captured via AR/VR devices), ensuring better attention correspondence between models and users. GazeXplain \cite{chen2024gazexplain} predicted visual scanpaths and generated natural language explanations for fixations using BLIP \cite{li2022blip}. However, its explanations remained semantic-level parsing, describing what is attended to without explicitly uncovering the underlying cognitive motivations behind attention allocations.
In contrast, LLada introduces a unified framework that jointly models spatial (\textit{where}), semantic (\textit{what}), and cognitive (\textit{why}) factors, enabling a more comprehensive and explainable approach to driver attention modeling. To the best of our knowledge, this is the first attempt to model underlying cognitive motivations of attention using MLLMs.

%% file: sec/3_dataset.tex
\section{\textbf{\textrm{W$^{3}$DA Dataset}}}

% 名字是 W^3DA  
% 我们的核心在于：统一现有的DR(eye)VE,LBW,BDDA,DADA主流驾驶注意力数据集（其中包含了各种场景：城市/高速/农村，正常驾驶/紧急制动/交通事故，白天/晚上），我们注释了驾驶注意区域对应的语义并进一步结合场景特有上下文解释驾驶注意背后的成因和动机，促进领域对驾驶注意的更全面可解释的建模和理解（从where到3W（where、what、why））。
% 需要简单讨论下 为什么不使用dino等目标检测模型（原因在于驾驶注意区域不一定对应一个特定物体（例如道路消失点等））
% 需要一个表格 总结所用到的4个数据集
% 可以多展现 我们数据的多样性：各种天气、各种场景、正常驾驶/紧急制动/交通事故

% 正常驾驶
% DReyeVE
% LBW
% 突发紧要情况 （紧急制动、交通事故）
% BDDA
% DADA

% (1) 简要的intro
% 主打亮点：whatwhy注释 + 多数据集多场景
% 我们的目标是实现对人在驾驶任务驱动下的视觉注意力机制的全面理解，突破现有范式通过生成热图仅仅实现预测注意空间位置的局限，从像素空间跨越到语义空间和认知知识空间。一个主要挑战就在于根本没有合适的数据集进行模型训练和评估。
% 为了解决这一问题，我们提出一种场景上下文感知的注意注释管道为现有的仅有注意区域(where)的驾驶注意数据提供高质量的注释目标（what）和注意成因（why）的注释，构建我们的W3AD数据集。同时，在W3AD中整合了多种主流的驾驶注意数据来源，包括正常驾驶DR(eye)VE、LBW，突发紧要场景的BDDA和交通事故的DADA-2000。这确保了W3AD的情景多样性，允许模型在多个数据集上进行协同训练，以增强其通用性。

\begin{figure*}[t]
\centering
	\includegraphics[trim=15 20 15 15, clip=true, width=0.90\textwidth]{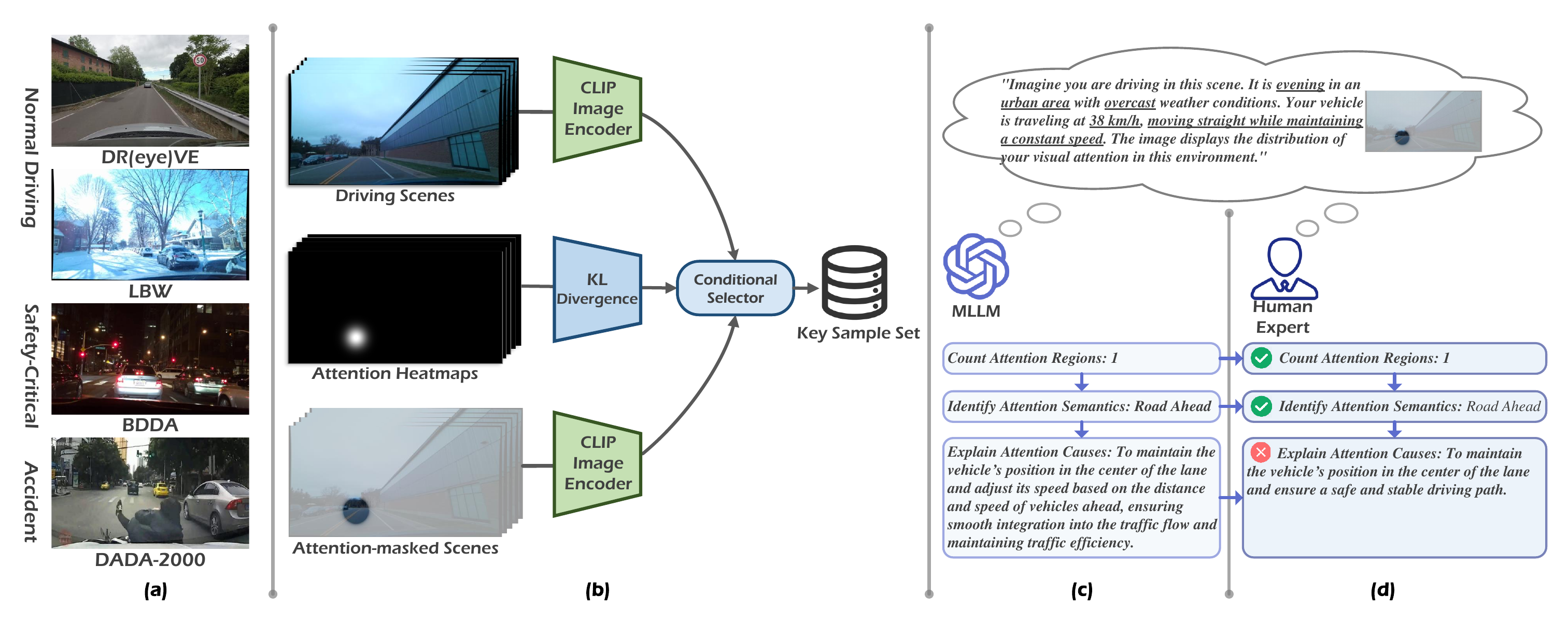}
	\caption{The W³DA annotation pipeline. (a) W³DA integrates multiple driver attention benchmarks, covering normal driving, safety-critical scenarios, and accidents. (b) To reduce annotation redundancy in scenarios with stable attention patterns (e.g., cruising, following another vehicle), we propose a key sample selection method, leveraging semantic scene similarity, spatial attention divergence, and semantic attention similarity to identify key moments of significant attention shifts. (c) We employ visual and contextual prompts with a structured chain-of-thought to guide an advanced MLLM in generating preliminary context-aware annotations. (d) Finally, human experts validate and refine these annotations to ensure accuracy, contextual relevance, and alignment with real driving scenarios.} 
	\label{fig2}
\end{figure*}

% W3da的数据标注过程。（a）W3DA整合了现有的多种驾驶注意benchmark，包括normal driving，safety-critical和accident。（b）因为很多场景（例如cruising，following another vehicle等），驾驶注意呈现相似不变性，为减少标注及训练冗余，我们提出一种key sample selection method based on attention dynamics（包括考虑场景的语义相似性、注意力空间位置相似性、注意力区域语义相似性等）来获取identifying key moments indicative of significant attention shifts by evaluating perceptual and attention-related changes between consecutive samples, ensuring the annotations capture high-information moments。（c）我们使用visual and contextual prompts with a clever chain-of-thought to guide a strong closed-source MLLM in generating preliminary annotations that are contextually relevant. (4) 最后，we enlist human experts, who, provided with the same inputs as the MLLM, carefully review and refine the model’s outputs based on their domain expertise. This ensures that the final annotations are accurate, reasonable, and closely aligned with the contextual information of the driving scenario. 

Existing driving attention datasets primarily predict where drivers look via spatial heatmaps. However, they lack semantic and cognitive understanding, limiting their ability to comprehensively model task-driven visual attention in driving. To bridge this gap, we move beyond spatial attention prediction to a deeper, more explainable understanding of driver attention. A key challenge is the absence of a dataset that provides semantic parsing (\textit{what}) and cognitive reasoning (\textit{why}) for driver attention.

To address this, we introduce W³DA, a semantic and reasoning-aware driving attention dataset that enhances conventional \textit{where} annotations with high-quality \textit{what} and \textit{why} labels. 
As shown in Fig. \ref{fig2}(a), W³DA unifies multiple mainstream driving attention datasets, integrating DR(eye)VE \cite{Palazzi2019dreyeve} and LBW \cite{LBW} for normal driving, BDDA \cite{xia2019predicting} for critical situations, and DADA-2000 \cite{fang2019dada,fang2022DADA} for traffic accidents. This diverse data composition allows for more robust cross-domain training, improving model generalization and real-world applicability.
Furthermore, as depicted in Fig. \ref{fig2}(b-d), we introduce an efficient and highly reliable data annotation pipeline that assigns semantic labels to attention targets (\textit{what}) and provides context-aware cognitive reasoning for attention causes (\textit{why}), enabling a more comprehensive understanding of driver attention.

\subsection{Data Annotation Pipeline}
% 使用MLLM进行标注：自动省时、但不一定准确
% 人工专家注释：费时、但相对准确
% 我们的方案 半自动化的MLLM-人工混合数据生成：

% 先介绍存在难点，再引入我们的方法
% 想要为实现可解释驾驶注意的数据生成存在几大固有难点。
% （1）驾驶注意数据繁多，全部逐帧进行标注将会有巨大的时间和财力成本，而通常在大多数时刻下驾驶员的注意力不会发生特别大的转移（例如驾驶员直行或跟车或（你再帮我想想具体例子））。
% （2）人工进行标注耗时耗力，而是用MLLM则是可行的替代思路，可以节省，但又存在不准确、不合理等问题。
% 我们的解决方案：引入一种Context-Aware的半自动人机协作的数据生成pineline。

Creating a large-scale, high-quality dataset for explainable driver attention prediction presents several inherent challenges. First, while vast amounts of driver attention data exist, many driving scenarios—such as cruising, highway driving, or following another vehicle—exhibit stable attention patterns, with drivers primarily fixating on the preceding vehicle or the road’s vanishing point. In such cases, attention shifts are minimal, making frame-by-frame annotation inefficient and resource-intensive.
Secondly, manual annotation by human experts incurs substantial time and financial costs. While leveraging powerful MLLMs for automated annotation is a promising alternative, MLLMs are prone to hallucinations, often generating erroneous or contextually inappropriate annotations that compromise dataset reliability.
To address these challenges, we introduce an efficient and highly reliable semi-automatic pipeline for dataset annotation.

% 对于第一个挑战，我们提出一种注意语义-aware的关键样本选择方法，我们不进行冗余的逐帧注释，而是通过计算帧之间的感知差异和注意认知差异来选择可能反映重大注意力转移的关键时刻，从而进行高信息量的注释。具体的，我们将V∈RT×H×W×3表示为原本的驾驶视频，我们计算一个关键样本集V‘∈RT’×H×W×3足够稀疏但全面覆盖驾驶视频的关键事件。我们考虑3个重要指标：（1）驾驶场景感知语义相似性 （2）驾驶员注意的空间位置相似性（3）驾驶员注意的语义相似性。在实践中，我们用V的第一帧初始化关键帧集V‘。对于V中的每一帧，我们计算其与V‘中最新关键帧的3个指标。若有一个指标低于预定义的阈值，我们将帧视为关键帧并将其添加到V‘。如果没有，则将跳过帧作为冗余。
For the first challenge, we introduce an attention-aware key sample selection strategy that avoids redundant frame-by-frame annotation. This approach prioritizes identifying key moments indicative of significant attention shifts by evaluating perceptual and attention-related changes between consecutive samples, ensuring the annotations capture high-information moments. 
Our selection process is grounded in three key criteria: (1) \textbf{semantic similarity of driving scenes}, capturing perceptual variations in the environment; 
(2) \textbf{spatial divergence of driver attention}, capturing variations in the spatial distribution of attended regions; and (3) \textbf{semantic similarity of driver attention}, reflecting contextual shifts in attended regions.
Specifically, criteria (1) and (3) are computed using the CLS token from the CLIP-Large image encoder \cite{radford2021learning}, while (2) is evaluated via KL divergence between attention heatmaps.
To implement this, we initialize the key sample set \( K \) with the first sample from the original driving video \( V \in \mathbb{R}^{T \times H \times W \times 3} \). For each subsequent sample in \( V \), we evaluate these three metrics relative to the most recent key sample in \( K \). 
A sample is added to \( K \) if the KL divergence (criterion 2) exceeds a predefined threshold or if the semantic similarity metrics (criteria 1 and 3) fall below their respective thresholds. Otherwise, it is considered redundant and excluded from further consideration.

% 是否要增加伪代码 （更多细节见补充材料）

% 对于第二个问题，我们引入Context-Aware的半自动人机协作...
% 为了解决第二个挑战，我们先通过设计巧妙地视觉和上下文的提示以及从简到难的思维链，来引导强大的闭源MLLM生成初步的可用的/可接受的/注释，再进一步由人类专家确定最终准确且可靠的注释。具体的，我们为MLLM提供充足的驾驶行为、天气条件和场景类别等上下文提示，促使MLLM生成与当前情景高度关联的what和why注释，并通过一个简单而有效的灰度蒙版将注意力叠加在图像上，帮助MLLM的注意力聚焦在驾驶注意区域。随后，还通过一个巧妙地思维链来引导MLLM从简到难的解析驾驶注意的语义和背后潜在成因。该思维链要求mllm首先确定场景中有多少个注意力区域，然后解析每个区域是什么（what），再深入挖掘问每一注意区域的成因（why））。我们在预实验中构建了多种方案变式，最后证明了这一套pineline的应用的有效性。
% 尽管MLLM已经具备强大的视觉推理能力，但依然可能产生幻觉。我们聘请了人类专家，在给人类专家与MLLM相同输入的情况下，根据human expertise来仔细审查并修正MLLM的输出结果，以得到最终准确、合理、紧密符合场景上下文的注释。
% efficient and highly reliable semi-automatic pipeline for dataset generation.

For the second challenge, we propose an effective approach that integrates visual and contextual prompts with a clever chain-of-thought strategy to guide an advanced MLLM in generating preliminary annotations that are both useful and contextually relevant. Specifically, we provide the MLLM with comprehensive contextual cues, including driving behavior, weather conditions, and scene categories, encouraging it to generate annotations aligned with the current scenario. Additionally, we overlay attention maps onto images using a simple grayscale mask, directing the MLLM's focus toward the relevant regions where driver attention is concentrated.
We further employ a chain-of-thought to guide the MLLM through a progressive analysis of the semantics and underlying causes of driver attention. This process prompts the MLLM first to determine the number of attention regions within the scene, then identify the content within each region (answering \textit{what}), and finally explore the causal factors behind the attention allocation (answering \textit{why}). In practice, we utilize the Qwen-VL-Max API as our MLLM to implement this approach.

Despite the MLLM’s strong visual reasoning capabilities, hallucinations remain a potential issue. 
To address this, we involve human experts to review and refine the model’s outputs based on their domain expertise, ensuring the final annotations are accurate, reasonable, and contextually aligned with the driving scenario.

% 更多细节（如prompt模板）将放在补充材料中。

\subsection{Dataset Statistics}
 % 总视频数量、总时间长度、总标注数量
 % 词云（what）（why）、平均注意力图

% 最终，我们的数据集包含从3,548视频场景中提取到的69980个关键帧，其中正常驾驶场景包含22839帧，危急场景包括22950帧，事故场景包含24191帧；涵盖了丰富的天气（sunny,rainy,cloudy,overcast,snowy,foggy）、地理位置（urban，rural，highway,suburban，mountain, tunnel）和时间（上午、下午、晚上），平均的what文本注释长度为5.40个单词，平均why文本注释长度为24.32个的单词。
% W3DA unlock a deeper understanding of driver attention mechanisms, providing valuable insights to the research community.
% 更多数据统计分析请见补充材料。

Building on our pipeline, W³DA comprises 69,980 key samples extracted from 3,548 video scenes, covering diverse driving conditions. Specifically, it includes 22,839 samples from normal driving, 22,950 samples from critical situations, and 24,191 samples from accidents. W³DA spans a variety of weather conditions (sunny, rainy, cloudy, overcast, snowy, foggy), geographic locations (urban, rural, highway, suburban, mountain, tunnel), and time periods (morning, afternoon, night).
Furthermore, W³DA provides rich textual annotations: the average length of \textit{what} annotations is 5.40 words, while the average length of \textit{why} annotations is 24.32 words, offering fine-grained semantic and cognitive insights into driver attention. 
To maintain a rigorous experimental setup, we adhere to the original training-validation-test splits of the four source datasets, ensuring no data leakage.
W³DA advances driver attention modeling, offering valuable insights for both academia and industry. 
More details can be found in the supplementary material.

%% file: sec/4_method.tex
\section{Explainable Driver Attention Prediction}
% Large Language-driven Driving Attention
% LLADA

\subsection{Problem Definition}
% 任务意义
% 任务形式化描述

% Our objective is to simultaneously predict pixel-level attention regions (where), provide semantic-level interpretations (what), and perform cognitive reasoning to explain the underlying causes of attention allocation (why), enables a more comprehensive and explainable modeling of driver attention. 
% 公式上，给定一个驾驶场景，Explainable Driver Attention Prediction的目标在于生成attention map，以及对应的一组语义解释和一组潜在成因解释。

The goal of Explainable Driver Attention Prediction is to predict \textbf{where a driver is likely to attend, what regions are attended, and why attention is allocated}. This task goes beyond conventional attention prediction by unifying spatial, semantic, and causal reasoning, enabling a more explainable and cognition-driven understanding of driver attention.
Formally, given a driving scene represented as an image $\mathbf{I} \in \mathbb{R}^{H \times W \times C}$ with contextual information $\mathbf{C}$ (e.g., weather, road conditions, driving behavior), the goal is to estimate a pixel-wise attention map $\mathbf{A} \in \mathbb{R}^{H \times W}$, a set of semantic descriptions of attended regions $\mathcal{S} = \{s_i\}_{i=1}^{N}$, and a set of causal explanations for attention allocation $\mathcal{E} = \{e_i\}_{i=1}^{M}$, where $s_i$ denotes the semantic label of the $i$-th attended region, and $e_i$ represents the explanation behind the attention allocated to the $i$-th region.

\begin{figure*}[t]
        \centering
	\includegraphics[trim=15 15 15 15, clip=true, width=0.87\textwidth]{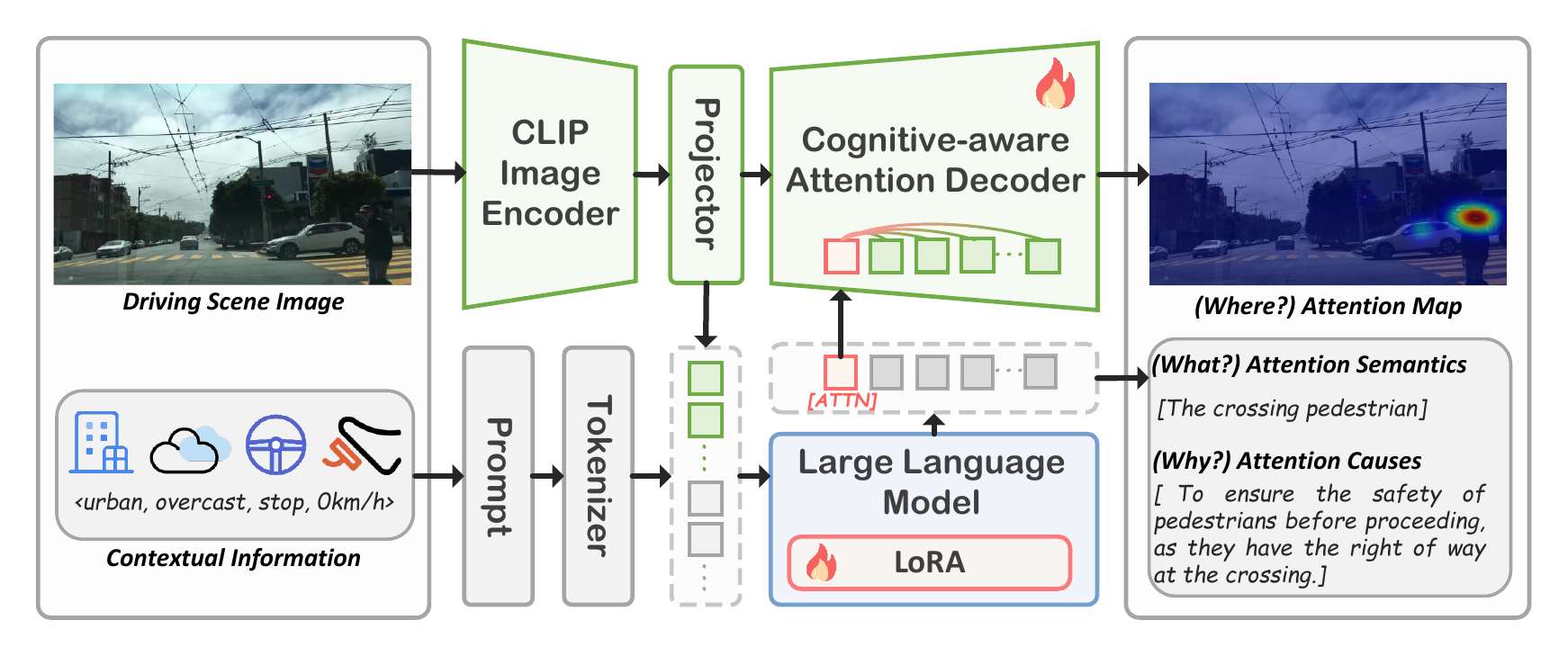}
        \caption{Overall architecture of LLada. Given a driving scene represented as an image with contextual information, the image is processed by the visual encoder (CLIP encoder and projector) to extract visual tokens, while the contextual information, along with prompt sequences, is tokenized by the LLM tokenizer. The visual and text tokens are then fed into the LLM. To adapt the LLM for explainable attention prediction tasks, a special attention token $[\text{ATTN}]$ is introduced to encode high-level cognitive cues. Finally, $[\text{ATTN}]$ performs cross-attention with the visual tokens, generating the attention map (answering \textit{Where?}) through cognitive-aware attention decoder, while the LLM produces the semantic descriptions (answering \textit{What?}) and causal explanations (answering \textit{Why?}) of attended regions.}
	\label{fig3}
\end{figure*}

\subsection{LLada Architecture}
% 近年，以大规模语言模型（LLM）为基础的多模态大模型在许多视觉任务中取得了显著进展，跨越image-level captioning、object-level detection和pixel-level segmentation。特别是，LLM的强大推理能力被成功引入视觉任务中，推动了视觉认知能力的飞跃。这些模型在复杂视觉认知任务中展现出了优越的表现，尤其是在理解图像内容、生成文本描述以及多模态任务的融合上。遗憾的是，在当前在注意力/显著性预测领域，据我们所知尚未有研究探讨如何基于LLM构建具有推理能力的注意力/显著性计算模型，严重限制了对注意力机制的背后成因的深层建模和理解。

%Recent advancements in multimodal large language models (MLLMs) have demonstrated remarkable progress across various vision tasks, spanning image-level captioning, object-level detection, and pixel-level segmentation. Notably, the powerful reasoning capabilities of large language models (LLMs) have been successfully introduced into vision tasks, leading to a paradigm shift in visual cognition. 
%\textcolor{red}{Despite these advances, LLM-driven attention/saliency prediction remains largely unexplored. To the best of our knowledge, no prior work has leveraged LLMs for attention modeling with reasoning capabilities, limiting a deeper understanding of visual attention, particularly its underlying cognitive motivations. To bridge this gap, we propose \textbf{LLada}, a \textbf{L}arge \textbf{La}nguage model-driven framework for \textbf{d}river \textbf{a}ttention prediction, as shown in Fig. \ref{fig3}.}

%\noindent \textbf{Framework Overview.} 
LLada comprises four core components: 1) a pretrained visual encoder $\mathcal{F}_\text{vis}$, 2) a large language model $\mathcal{F}_{\text{LLM}}$, 3) a special attention token $[\text{ATTN}]$, and 4) a cognitive-aware attention decoder $\mathcal{F}_{\text{dec}}$, as illustrated in Fig. \ref{fig3}. 
Given a driving scene, the visual input $\mathbf{I}$ and contextual information $\mathbf{C}$ are first processed by the pretrained visual encoder $\mathcal{F}_\text{vis}$ and the LLM $\mathcal{F}_\text{LLM}$, respectively. Then, the attention token $[\text{ATTN}]$ encodes high-level cognitive cues within $\mathcal{F}_\text{LLM}$ and enables interaction between the language and vision modalities through the attention decoder $\mathcal{F}_{\text{dec}}$. Finally, $\mathcal{F}_{\text{dec}}$ generates the spatial attention map $\mathbf{A}$, while the $\mathcal{F}_{\text{LLM}}$ outputs the semantic descriptions $\mathcal{S}$ and causal explanations $\mathcal{E}$. Next, we provide a detailed description of each component.

%\noindent \textbf{Visual Encoder and LLM}
%Visual Encoder and LLM，我们遵循现有的MLLM设置，一个基于解码器的语言模型LLM，以根据用户的输入自动回归生成文本响应，一个视觉编码器FV1，以从输入的图像中提取特征，以及一个线性投影仪ϕ，以对齐图像和文本模式之间的表示。具体来说，使用了具有CLIP-ViT-L/14 [55]和Vicuna-7B/13B [6]的预训练的LLaVA [39]变体。给定一个输入图像ximg，视觉编码器FV1首先将其编码为图像特征，然后投影仪ϕ将这些特征映射到LLM输入空间中的视觉标记嵌入中：himg =ϕ（FV1（ximg）），
%输入图像himg通常调整为h×w×3，图像标记himg∈Rnimg×与语言模态对齐。对于CLIP-ViT-L/14，使用= 224的输入图像用14的补丁大小进行编码，因此标记的长度为=/142=256和Vicuna-的LLM尺寸d分别为4096和5120
%与输入图像一起，描述要场景上下文的文本指令被LLM标记发生器T： htxt = T（xtxt）标记为文本标记。图像标记和文本标记被连接在一起，然后在预先一系列固定的提示h提示后将其输入LLM。输出令牌嵌入˜ytxt是自动回归生成的：˜ytxt=FLLM（[h提示符||himg||htxt]），(3)，
%其中||是连接操作。文本响应从˜ytxt中通过应用线性分类器来预测词汇表中的下一个单词。

\noindent \textbf{Visual Encoder.}
We utilize a visual encoder consisting of the pretrained vision foundation model CLIP-ViT-L \cite{radford2021learning} and a linear projector $\phi$, following the setup of LLaVA \cite{liu2023visual}. Given the visual input $\mathbf{I}$, CLIP extracts image features, which are then projected by $\phi$ into the visual token space, producing embeddings $\mathbf{h}_{\text{vis}} \in \mathbb{R}^{N_{\text{}} \times d}$ for further processing by $\mathcal{F}_{\text{LLM}}$.

\noindent \textbf{Large Language Model.}
The driving scene context $\mathbf{C}$ is tokenized into context tokens by the LLM tokenizer $\mathcal{T}$: $\mathbf{h}_{\text{con}} = \mathcal{T}(\mathbf{C})$. The visual tokens $\mathbf{h}_{\text{vis}}$ and context tokens $\mathbf{h}_{\text{con}}$ are then concatenated with a fixed prompt sequence $\mathbf{h}_{\text{prompt}}$ and fed into the LLM, denoted as $\mathcal{F}_{\text{LLM}}$.
The output token embeddings $\tilde{\mathbf{y}}_{\text{txt}}$ are generated as:
\begin{align}
    \tilde{\mathbf{y}}_{\text{txt}} = \mathcal{F}_{\text{LLM}}([\mathbf{h}_{\text{prompt}}, \mathbf{h}_{\text{vis}}, \mathbf{h}_{\text{con}}]),
\end{align}
where $[\cdot]$ denotes concatenation. The final text responses are generated by applying a linear classifier to $\tilde{\mathbf{y}}_{\text{txt}}$ for next-token prediction. During attention prediction, these responses include semantic descriptions of the attended regions $\mathcal{S}$ and causal explanations for attention allocation $\mathcal{E}$. In practice, we use Vicuna-7B \cite{vicuna2023} as the $\mathcal{F}_{\text{LLM}}$.

\noindent \textbf{Attention Token.}
% 为了将LLM适配于注意力预测任务，我们使用一个特殊的新的注意力标记来扩展原始的LLM词汇表，即[ATTN],用于encodes high-level cognitive cues within LLM and facilitates interaction between the language and vision modalities through the decoder.当LLM执行注意力预测任务，输出的ˆytxt将包含一个<ATTN>令牌，我们选择[ATTN]标记˜ytxt[ATTN]的输出嵌入，并通过MLP投影仪ψ：hattn=ψ（˜ytxt[ATTN]）将其投影到attention decoedr的空间中。因此，attention decoder可以从令牌hattn中解码注意力认知信息。
To adapt the LLM for attention prediction, we introduce a special attention token, $[\text{ATTN}]$, extending the original LLM vocabulary. This token encodes high-level cognitive cues within the LLM and  facilitates cross-modal interaction between language and vision via $\mathcal{F}_{\text{dec}}$. During attention prediction, the output token sequence $\tilde{\mathbf{y}}_{\text{txt}}$ includes the $[\text{ATTN}]$ token. We extract its corresponding embedding $\tilde{\mathbf{y}}_{\text{txt}}[\text{ATTN}]$, which is then projected into the $\mathcal{F}_{\text{dec}}$ space via an MLP projector $\psi$:
\begin{align}
    \mathbf{h}_{\text{attn}} = \psi(\tilde{\mathbf{y}}_{\text{txt}}[\text{ATTN}]).
\end{align}
This enables $\mathcal{F}_{\text{dec}}$ to decode high-level cognitive attention information from $\mathbf{h}_{\text{attn}}$.

\noindent \textbf{Cognitive-aware Attention Decoder.}
The attention decoder $\mathcal{F}_{\text{dec}}$ facilitates the interaction between the attention token embedding $\mathbf{h}_{\text{attn}}$ and the visual features $\mathbf{h}_{\text{vis}}$, decoding context-aware cognitive information to generate the pixel-wise attention map $\mathbf{A} \in \mathbb{R}^{H \times W}$. Specifically, we first introduce a cross-attention mechanism to guide the decoder's awareness of high-level cognitive cues from $\mathbf{h}_{\text{attn}}$ and derive cognitively-driven visual features $\mathbf{h}_{\text{vis}}'$:
\begin{align}
\mathbf{h}_{\text{dec}}' = \mathbf{h}_{\text{vis}} + \text{Repeat}(CA(\mathbf{h}_{\text{attn}}, \mathbf{h}_{\text{vis}})),
\end{align}
where $CA(q, kv)$ denotes the cross-attention operation and $\text{Repeat}(\cdot)$ is a replication operation, which adds $CA(\mathbf{h}_{\text{attn}}, \mathbf{h}_{\text{vis}})$ to each token in $\mathbf{h}_{\text{vis}}$. 
Next, the $\mathbf{h}_{\text{vis}}'$ are reshaped from a 2D shape of $\frac{HW}{p^2} \times C$ to a standard 3D feature map of size $\frac{H}{p} \times \frac{W}{p} \times C$, where $C$ is the feature channel size and $p$ is the image patch size. Then, we apply a series of five $3 \times 3$ convolutional layers with batch normalization and ReLU activation to reduce the feature dimensionality to $\frac{H}{p} \times \frac{W}{p}$. Finally, a bilinear upsampling operation is applied to generate the attention map $\mathbf{A}$ at the full image size of $H \times W$.

% \textbf{} CLIP LLM
% TOKEN
% decoder

\noindent \textbf{Training Objectives.}
The LLada model is trained end-to-end using the attention map prediction loss $\mathcal{L}_{\text{map}}$ and the textual explanation generation loss $\mathcal{L}_{\text{txt}}$. The overall objective $\mathcal{L}$ is the weighted sum of these losses:
\begin{align}
\mathcal{L} = \lambda_{\text{map}} \mathcal{L}_{\text{map}} + \lambda_{\text{txt}} \mathcal{L}_{\text{txt}},
\end{align}
where $\lambda_{\text{map}}$ and $\lambda_{\text{txt}}$ are scaling factors. Specifically, the attention map loss $\mathcal{L}_{\text{map}}$ consists of binary cross-entropy (BCE) and Kullback-Leibler (KL) divergence, encouraging accurate pixel-wise attention maps (answering \textit{Where?}):
\begin{align}
\mathcal{L}_{\text{map}} = \lambda_{\text{bce}} \mathbf{BCE}(\hat{\mathbf{A}}, \mathbf{A}) + \lambda_{\text{kl}} \mathbf{KL}(\hat{\mathbf{A}}, \mathbf{A}),
\end{align}
where $\hat{\mathbf{A}}$ is the predicted attention map and $\mathbf{A}$ is the ground truth.
The textual explanation generation loss $\mathcal{L}_{\text{txt}}$ is formulated as the autoregressive cross-entropy (CE) loss, guiding the LLM to generate correct semantic descriptions (answering \textit{What?}) and causal explanations (answering \textit{Why?}) of the attended regions:
\begin{align}
\mathcal{L}_{\text{txt}} = \lambda_{\text{what}}\mathbf{CE}(\hat{\mathcal{S}}, \mathcal{S}) + \lambda_{\text{why}}\mathbf{CE}(\hat{\mathcal{E}}, \mathcal{E}),
\end{align}
where $\hat{\mathcal{S}}$ and $\hat{\mathcal{E}}$ are the predicted semantic descriptions and causal explanations, respectively. That $\mathcal{S}$ and $\mathcal{E}$ are the ground truth descriptions and explanations.

%% file: sec/5_experiment.tex
\section{Experiment}

% 大规模MLLM+人工
% 再加一个小规模 完全人工的测试集 （比如4*100个）

% 文本比较 （what和why单独分开）
% 用BLIP构建baseline （attn mask+BLIP // 局部裁剪+BLIP）
% 单独训练注意分支+原生LLAVA
% 人工主观评估
\subsection{Experimental Setting}
\noindent \textbf{Implementation Details.}
We train our model on 4 NVIDIA A100 GPUs, utilizing the DeepSpeed \cite{rasley2020deepspeed} engine for efficient optimization. The training process employs the AdamW \cite{loshchilov2017decoupled} optimizer with a learning rate of 0.0003, and WarmupDecayLR as the learning rate scheduler with 100 warmup iterations. 
For loss balancing, we set the scaling factors \(\mathcal{L}_{\text{map}}\), \(\mathcal{L}_{\text{txt}}\),  \(\mathcal{L}_{\text{bce}}\), \(\mathcal{L}_{\text{kl}}\), \(\mathcal{L}_{\text{what}}\), and \(\mathcal{L}_{\text{why}}\)  to 2, 1, 1, 0.1, 1, and 1, respectively.
Training is conducted with a batch size of 8 per device, using gradient accumulation with a step size of 5 to accommodate memory constraints.  
To retain the knowledge in the pre-trained LLM \(\mathcal{F}_{\text{LLM}}\), we apply LoRA \cite{hu2022lora} for parameter-efficient fine-tuning while keeping the visual encoder \(\mathcal{F}_{\text{vis}}\) entirely frozen. The cognitive-aware attention decoder \(\mathcal{F}_{\text{dec}}\) is trained from scratch.

\noindent \textbf{Evaluation Metrics.}
We follow the evaluation setup of prior driver attention prediction studies \cite{chen2023fblnet,fang2022DADA} and assess the spatial accuracy of generated attention maps using AUC\textsubscript{J}, AUC\textsubscript{B}, SIM, CC, KLdiv, and NSS. To evaluate the language quality of the generated \textit{what} and \textit{why} textual explanations, we adopt BLEU, METEOR, ROUGE, and CIDEr-R. 
%\textcolor{red}{Finally, we conduct a user study to assess the results based on human subjective perception.} 
This comprehensive evaluation allows us to effectively measure the model’s performance in spatial, semantic, and cognitive reasoning for driver attention.

% \subsection{Comparison methods \& Metrics}
%  Benchmarks and Baselines

\subsection{Results on W³DA}
% 表1：W³DA-分3个场景/4个数据集下的注意力评估
% 表2：生成文本评估（+几个baseline 参考gazexplain）

\begin{table*}[t]
\centering
%\caption{Comparison of Attention Map Prediction Performance. \textbf{Bold} and \underline{underline} show the best and second-best performances. † denotes multi-task models optimizing both Attention Map Prediction (answering ``where") and Textual Explanation Generation (answering ``what/why").}
\caption{Comparison of attention map prediction performance. \textbf{Bold} and \underline{underline} indicate the best and second-best results. † marks multi-task models jointly optimizing Attention Map Prediction (\textit{where}) and Textual Explanation Generation (\textit{what/why}).}
\label{tab1}
\resizebox{\textwidth}{!}{%
\begin{tabular}{l|cccccc|cccccc|cccccc}
\toprule
\multirow{2}{*}{Method} & \multicolumn{6}{c|}{Normal Driving} & \multicolumn{6}{c|}{Safety-Critical Situation} & \multicolumn{6}{c}{Traffic Accident} \\ \cmidrule(lr){2-7} \cmidrule(lr){8-13} \cmidrule(lr){14-19}
& KLdiv↓ & CC↑ & SIM↑ & AUC\textsubscript{J}↑ & AUC\textsubscript{B}↑ & NSS↑ & KLdiv↓ & CC↑ & SIM↑ & AUC\textsubscript{J}↑ & AUC\textsubscript{B}↑ & NSS↑ & KLdiv↓ & CC↑ & SIM↑ & AUC\textsubscript{J}↑ & AUC\textsubscript{B}↑ & NSS↑ \\
\midrule
ITTI \cite{itti1998model}     & 3.216 & 0.093 & 0.080 & 0.676 & 0.665 & 0.662 & 2.807 & 0.049 & 0.119 & 0.618 & 0.613 & 0.410 & 3.339 & 0.033 & 0.073 & 0.595 & 0.600 & 0.266 \\
GBVS \cite{harel2006graph}     & 2.572 & 0.294 & 0.139 & 0.868 & 0.857 & 2.057 & 2.238 & 0.246 & 0.176 & 0.839 & 0.804 & 1.501 & 2.826 & 0.173 & 0.105 & 0.814 & 0.765 & 1.212 \\
DeepGaze I \cite{KummererTB14} & 3.103 & 0.207 & 0.080 & 0.885 & 0.546 & 1.254 & 2.550 & 0.237 & 0.137 & 0.874 & 0.595 & 1.297 & 3.060 & 0.206 & 0.082 & 0.883 & 0.673 & 1.412 \\
DeepGaze IIE \cite{Linardos_2021_ICCV} & 3.071 & 0.226 & 0.082 & 0.887 & 0.610 & 1.382 & 2.505 & 0.296 & 0.141 & 0.926 & 0.589 & 1.641 & 3.026 & 0.227 & 0.085 & 0.917 & 0.584 & 1.476 \\
MLnet \cite{dodge2018visual}    & 2.129 & 0.547 & \textbf{0.460} & 0.914 & 0.836 & 4.494 & 1.953 & 0.528 & \textbf{0.433} & 0.928 & 0.874 & 5.254 & 2.897 & 0.344 & \textbf{0.288} & 0.893 & 0.784 & 3.008 \\
CDNN \cite{deng2020CDNN}     & 2.614 & 0.465 & 0.394 & 0.887 & 0.790 & 3.687 & 2.646 & 0.401 & 0.350 & 0.885 & 0.767 & 3.831 & 3.568 & 0.283 & 0.254 & 0.851 & 0.714 & 2.446 \\
FBnet \cite{ding2021fbnet}    & 2.980 & 0.406 & 0.343 & 0.869 & 0.769 & 3.103 & 2.585 & 0.431 & 0.364 & 0.887 & 0.803 & 3.975 & 3.197 & 0.329 & 0.246 & 0.873 & 0.789 & 2.727 \\
ConvNeXt \cite{liu2022convnet} & 2.042 & \underline{0.570} & 0.412 & 0.916 & 0.848 & \underline{4.661} & 1.765 & \underline{0.567} & 0.413 & 0.938 & \underline{0.877} & \textbf{5.438} & 3.049 & 0.377 & 0.248 & 0.891 & 0.806 & \textbf{3.425} \\
ERFNet \cite{yang2024progressive}   & \underline{1.979} & 0.558 & 0.425 & \underline{0.923} & 0.840 & 4.304 & \underline{1.593} & 0.538 & 0.410 & \underline{0.942} & 0.868 & 5.201 & \underline{2.181} & \underline{0.391} & 0.253 & \underline{0.930} & \underline{0.846} & 3.035 \\
GazeXplain† \cite{chen2024gazexplain} & 2.578 & 0.477 & 0.389 & 0.857 & \underline{0.866} & 3.945 & 2.769 & 0.383 & 0.321 & 0.848 & 0.743 & 2.299 & 3.109 & 0.371 & 0.236 & 0.902 & 0.804 & 1.598 \\ \midrule 
LLada† (Ours) & \textbf{1.219} & \textbf{0.583} & \underline{0.436} & \textbf{0.952} & \textbf{0.908} & \textbf{5.376} & \textbf{1.230} & \textbf{0.579} & \underline{0.420} & \textbf{0.950} & \textbf{0.912} & \underline{5.271} & \textbf{1.927} & \textbf{0.396} & \underline{0.262} & \textbf{0.934} & \textbf{0.889} & \underline{3.216} \\
\bottomrule
\end{tabular}
}
\end{table*}

\begin{table*}[t]
\centering
\caption{Comparison of textual explanation generation performance. \textbf{Bold} and \underline{underline} show the best and second-best performances.}
\label{tab2}
\resizebox{\textwidth}{!}{%
\begin{tabular}{l|cccc|cccc|cccc}
\toprule
\multirow{2}{*}{Method} & \multicolumn{4}{c|}{Normal Driving} & \multicolumn{4}{c|}{Safety-Critical Situation} & \multicolumn{4}{c}{Traffic Accident} \\ \cmidrule(lr){2-5} \cmidrule(lr){6-9} \cmidrule(lr){10-13}
& BLEU↑ & METEOR↑ & ROUGE↑ & CIDEr-R↑ & BLEU↑ & METEOR↑ & ROUGE↑ & CIDEr-R↑ & BLEU↑ & METEOR↑ & ROUGE↑ & CIDEr-R↑ \\ 
\midrule
% ITTI \cite{itti1998model} + LLaVA \cite{liu2023visual} & 0.230 & 0.258 & 0.309 & 0.258 & 0.280 & 0.273 & 0.344 & 0.297 & 0.262 & 0.251 & 0.327 & 0.294 \\
DeepGaze I \cite{KummererTB14}  + LLaVA \cite{liu2023visual} & 0.281 & 0.274 &	0.357 &	0.287 &	0.130 & 0.178 &0.269 & 0.078 & 0.179 & 0.213 & 0.291 & 0.167 \\
DeepGaze IIE \cite{Linardos_2021_ICCV} + LLaVA \cite{liu2023visual} & 0.281 & 0.275 & 0.361 & 0.301 & 0.253 & 0.283 & 0.346 & 0.289 & 0.253 & 0.266 & 0.344 & 0.272 \\
MLnet \cite{dodge2018visual} + LLaVA \cite{liu2023visual} & 0.294 & 0.283 & 0.398 & 0.279 &	0.285 &	0.294 &	0.395 & 0.312 & 0.258 &	0.264 &	0.373 & 0.260 \\
CDNN \cite{deng2020CDNN} + LLaVA \cite{liu2023visual} & 0.299 & 0.287 & \underline{0.403}	& 0.283	& 0.285	& 0.294	& 0.397 & 0.312 & 0.263	& 0.270	& 0.377 & 0.265 \\
ConvNeXt \cite{liu2022convnet} + LLaVA \cite{liu2023visual} & 0.291 & 0.282 & 0.396  & 0.268 & 0.117 & 0.173 & 0.268  & 0.078 & 0.225 & 0.247 & 0.345 & 0.209 \\
ERFNet \cite{yang2024progressive} + LLaVA \cite{liu2023visual} & 0.294 & 0.283 & 0.396  & 0.274 & \underline{0.287} & \underline{0.297} & \underline{0.398}  & 0.307 & \underline{0.270} & \underline{0.272} & 0.375  & 0.258 \\
GazeXplain \cite{chen2024gazexplain} & \underline{0.307} & \underline{0.299} & 0.216  & \underline{0.422} & 0.185 & 0.290 & 0.367  & \underline{0.554} & 0.167 & 0.195 & \underline{0.438} & \underline{0.656} \\ \midrule 
LLada (Ours) & \textbf{0.436} & \textbf{0.360} & \textbf{0.582}  & \textbf{0.963} & \textbf{0.444} & \textbf{0.375} & \textbf{0.593} & \textbf{1.233} & \textbf{0.376} & \textbf{0.318} & \textbf{0.520} & \textbf{1.002} \\
\bottomrule
\end{tabular}
}
\end{table*}

%（回答where）
\noindent \textbf{Attention Map Prediction (Where).}
We first evaluate the attention map prediction quality on the W³DA dataset, comparing our proposed LLada with 10 state-of-the-art methods: ITTI \cite{itti1998model}, GBVS \cite{harel2006graph}, DeepGaze I \cite{KummererTB14}, DeepGaze IIE \cite{Linardos_2021_ICCV}, MLNet \cite{dodge2018visual}, CDNN \cite{deng2020CDNN}, FBNet \cite{ding2021fbnet}, ConvNeXt \cite{liu2022convnet}, ERFNet \cite{yang2024progressive}, and GazeXplain \cite{chen2024gazexplain}. ITTI and GBVS are early saliency models based on handcrafted features, while the others are learning-based approaches. DeepGaze I and DeepGaze IIE use official pre-trained models due to unavailable training codes. MLNet, CDNN, FBNet, ConvNeXt, and ERFNet are specialized attention prediction models optimized solely for this task, whereas GazeXplain and LLada jointly optimize attention map prediction and textual explanation generation. 
Table \ref{tab1} presents the performance comparison. LLada surpasses all SOTA methods in every scenario. In particular, LLada achieves substantial gains on the KLdiv,  AUC\textsubscript{J}, and AUC\textsubscript{B} metrics. For KLdiv, it outperforms the second-best method (ERFNet) by 38.40\%, 22.79\%, and 11.65\% in normal driving, safety-critical situations, and traffic accidents, respectively. 
Compared to GazeXplain, which jointly optimizes attention map prediction (\textit{where}) and textual explanation generation (\textit{what} \& \textit{why}), LLada consistently achieves superior performance across all metrics, demonstrating its effectiveness.

\noindent \textbf{Textual Explanation Generation (What \& Why).}
%（回答whatwhy）
% 随后，我们在W³DA上评估textual explanations生成的质量。我们将我们提出的LLada同多任务模型GazeXplain进行比较。此外，我们还构建了一些强大的基线进行比较，以进一步验证我们方法LLada的有效性。与我们方法端到端训练相反的，我们构建一种更常见思路的双阶段方法，一阶段是通过一个强大的注意力预测模型获取注意力图（就像之前所有传统的注意力预测方法一样）；再单独训练一个解释注意力图的MLLM（使用的是LLAVA），随后将将从一阶段获得的注意力图的灰度蒙版输入给MLLM，让其生成what和why的解释。实践上，我们从注意力图预测评估中选择了表现最好的ERFNet、ConvNeXt和经典方法ITTI、DeepGaze2e作为一阶段的注意力预测模型；我们选择和我们llada相同使用的MLLM，也就是LLAVA（CLIP-ViT-L+Vicuna-7B），进行重新训练（输入直接是注意力图，输出是what和why的文本解释）。
% 表二展示了Textual Explanation Generation的结果，我们提出的LLada在所有场景所有指标上均取得一致性的优异表现，超越了多任务模型GazeXplain和强大的双阶段基线。
We further evaluate the quality of textual explanations on the W³DA dataset, comparing our proposed LLada with the multi-task model GazeXplain. To ensure a comprehensive evaluation, we also establish strong baseline methods to validate LLada's effectiveness.
Unlike our end-to-end approach, these baselines adopt a two-stage paradigm: first, a conventional attention prediction model generates an attention map; second, an MLLM interprets the attention map to generate textual explanations for \textit{what} and \textit{why}.
Specifically, we select DeepGaze I, DeepGaze IIE, MLnet, CDNN, ConvNeXt, and ERFNet as first-stage predictors.
For the second stage, we fine-tune LLaVA \cite{liu2023visual} (CLIP-ViT-L \cite{radford2021learning} + Vicuna-7B \cite{vicuna2023}) via LoRA \cite{hu2022lora}, using the same MLLM as LLada to process grayscale attention maps for textual explanation generation.
As shown in Table \ref{tab2}, LLada consistently outperforms all baselines across scenarios and metrics, surpassing both GazeXplain and all two-stage baselines, which demonstrates its superior cognitive reasoning capabilities in driver attention analysis. 
The results from Table \ref{tab1} (Attention Map Prediction) and Table \ref{tab2} (Textual Explanation Generation) collectively confirm that our LLada surpasses all previous methods, showcasing its capability for a deeper understanding of driver attention mechanisms.

%Compared with Conventional domain-specific models
\subsection{Results on Conventional Attention Prediction}
% 表2-5：4个数据集全集下的注意力评估
% Results on 独立数据集？ 验证跨域训练的优异
% 跨数据集统一训练的作用
To evaluate the effectiveness of LLada trained on the cross-domain dataset W³DA, we compare it with domain-specific attention prediction models fully trained on DR(eye)VE \cite{Palazzi2019dreyeve}, BDDA \cite{xia2019predicting}, and DADA \cite{fang2019dada,fang2022DADA}. Performance is evaluated on the complete test sets of these datasets. 
As shown in Table \ref{tab_crossdataset}, LLada demonstrates competitive performance, particularly in terms of KLdiv, outperforming the second-best domain-specific models by 29.81\%, 20.69\%, and 5.49\% on the DR(eye)VE, BDDA, and DADA test sets, respectively.
Additionally, GazeXplain, trained with cross-domain multi-task learning on W³DA, also achieves promising results.
Notably, W³DA provides 39,642 keyframes for training, while domain-specific models use 28,632 (DR(eye)VE), 26,325 (BDDA), and 33,939 (DADA) frames. Despite the cross-domain setting, training LLada on W³DA does not significantly increase cost, highlighting its strong generalization across domains and the efficiency of key-frame selection in W³DA.

\begin{table*}[ht]
    \centering
    \caption{Comparison with domain-specific models for conventional attention map prediction. }%\textbf{Bold} and \underline{underline} show the best and second-best performances.}
    % \renewcommand{\arraystretch}{0.75}  % 压缩行高
    % ============ 子表 (a) ============
    %\vspace{0pt}
    \begin{subtable}[t]{0.3\textwidth}
        \centering
        \small \caption{DR(eye)VE test set.}
        % 方法1: 直接缩小字体
        \scriptsize
        % 方法2: 或者用 \scalebox 全局缩放
        \scalebox{0.78}{
        \begin{tabular}{@{}c l c c@{}}
        \toprule
        Training & Method & KLdiv↓ & CC↑ \\
        \midrule
        \multirow{8}{*}{DR(eye)VE}  
         & MLNet \cite{dodge2018visual}        & 2.00  & 0.44 \\
         & RMDN \cite{bazzani2017recurrent}      & 1.77  & 0.41 \\
         & DR(eye)VE \cite{Palazzi2019dreyeve}   & 1.40  & 0.56 \\
         & HWS \cite{xia2019predicting}          & 1.72  & 0.51 \\
         & SCAFNet \cite{fang2022DADA}      & \underline{1.35}  & \underline{0.59} \\
         & DVAM \cite{deng2023driving}         & 1.38  & 0.57 \\
         & TransConvNet \cite{xu2024transconvnet} & 1.37  & 0.58 \\
         & MTSF \cite{jin2024mtsf}         & 1.72  & 0.56 \\
        \midrule
        \multirow{2}{*}{W³DA}
         & GazeXplain    & 1.48  & 0.49 \\
         & LLada     & \textbf{1.04}  & \textbf{0.67} \\
        \bottomrule
        \end{tabular}
        }
    \end{subtable}
    %\hspace{2mm} % 控制间距
    \hfill
    % ============ 子表 (c) ============
    \begin{subtable}[t]{0.35\textwidth}
        \centering
        \caption{BDDA test set.}
        % 方法1: 直接缩小字体
        \scriptsize
        % 方法2: 或者用 \scalebox 全局缩放
        \scalebox{0.78}{
        \begin{tabular}{@{}c l c c c@{}}
\toprule
Training & Method & KLdiv↓ & CC↑ & SIM↑ \\
\midrule
\multirow{9}{*}{BDDA}  
 & HWS \cite{xia2019predicting}   & 2.07  & 0.48 & 0.35\\
 & U2NET \cite{qin2020u2}        & 1.47  & 0.56 & 0.36\\
 & MINET \cite{pang2020multi}        & 10.50 & 0.49 & 0.36\\
 & DRIVE \cite{bao2021drive}        & 13.83 & 0.32 & 0.26\\
 & SCAFNet \cite{fang2022DADA}         & 1.48  & 0.56 & \underline{0.40}\\
 & DBNET \cite{tian2022driving}        & 1.85  & 0.56 & 0.38\\
 & FBLNet \cite{chen2023fblnet}       & \underline{1.40}  & \textbf{0.64} & \textbf{0.47}\\
 & MTSF \cite{jin2024mtsf}         & 1.61  & 0.51 & 0.36\\
\midrule
\multirow{2}{*}{W³DA}  
 & GazeXplain    & 2.85  & 0.38 & 0.32\\
 & LLada     & \textbf{1.16}  & \underline{0.60} & \textbf{0.47}\\
\bottomrule
\end{tabular}}
    \end{subtable}
    \hfill
    % ============ 子表 (d) ============
    \begin{subtable}[t]{0.3\textwidth}
        \centering
        \caption{DADA test set.}
        % 方法1: 直接缩小字体
        \scriptsize
        % 方法2: 或者用 \scalebox 全局缩放
        \scalebox{0.78}{
\begin{tabular}{@{}c l c c c@{}}
\toprule
Training & Method & KLdiv↓ & CC↑ & SIM↑   \\
\midrule
\multirow{8}{*}{ DADA}  
 & HWS \cite{xia2019predicting}        & 2.77  & 0.33 & 0.22 \\
 & U2NET \cite{qin2020u2}         & \underline{1.86}  & 0.47 & 0.30\\
 & MINET  \cite{pang2020multi}  & 9.99  & 0.39 & 0.30\\
 & DRIVE \cite{bao2021drive}   & 4.03  & 0.37 & 0.25\\
 & SCAFNet  \cite{fang2022DADA}     & 2.17  & \underline{0.48} & 0.34\\
 & DBNET  \cite{tian2022driving}        & 2.77  & 0.40 & 0.26\\
 & PGNET \cite{xie2022pyramid} & 5.28  & 0.46 & \textbf{0.37}\\
 & FBLNet \cite{chen2023fblnet}          & 1.92  & \textbf{0.50} & 0.33\\
\midrule
\multirow{2}{*}{W³DA}  
 & GazeXplain    & 1.93  & 0.46 & 0.29\\
 & our LLada     & \textbf{1.82}  & \underline{0.48} & \underline{0.36}\\
\bottomrule
\end{tabular}}
    \end{subtable}
    \label{tab_crossdataset}
\end{table*}

\subsection{Qualitative Analyses}

\begin{figure*}[t]
        \centering
	\includegraphics[trim=15 18 15 18, clip=true, width=0.90\textwidth]{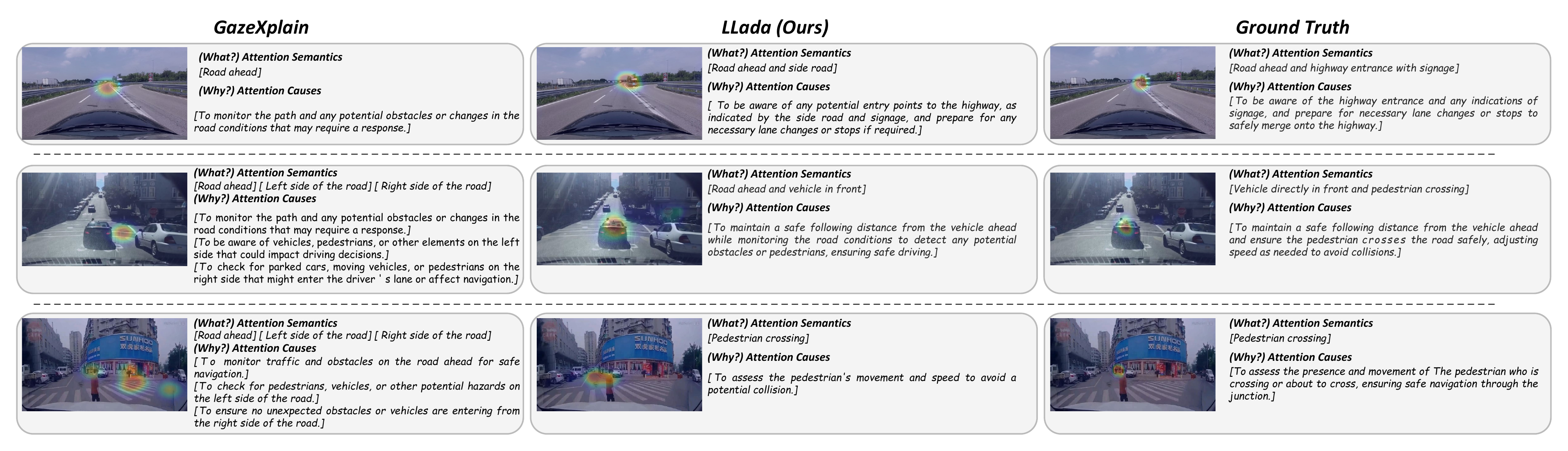}
        \caption{Quantitative examples. More in supplementary materials.}
	\label{fig4}
\end{figure*}

We quantitatively compare our LLada with the latest GazeXplain \cite{chen2024gazexplain} on explainable driver attention prediction, as shown in Fig. \ref{fig4}. LLada's pixel-wise attention map and textual explanations align more closely with the Ground Truth, capturing attention mechanisms across spatial, semantic, and cognitive levels. In complex scenarios, such as busy traffic flow (Row 2) and pedestrian interactions (Row 3), LLada outperforms GazeXplain. LLada correctly focuses on the pedestrian in front of the vehicle (Row 3) and provides context-aware, task-driven explanations like ``assessing the pedestrian’s movement and speed to avoid a potential collision." In contrast, GazeXplain misses this key focus and offers less relevant, scene-independent explanations. See supplementary materials for more examples.

% 更多可视化例子，请见补充材料

\subsection{Ablation Study}

\begin{figure}[t]
        \centering
	\includegraphics[trim=5 5 5 5, clip=true, width=0.41\textwidth]{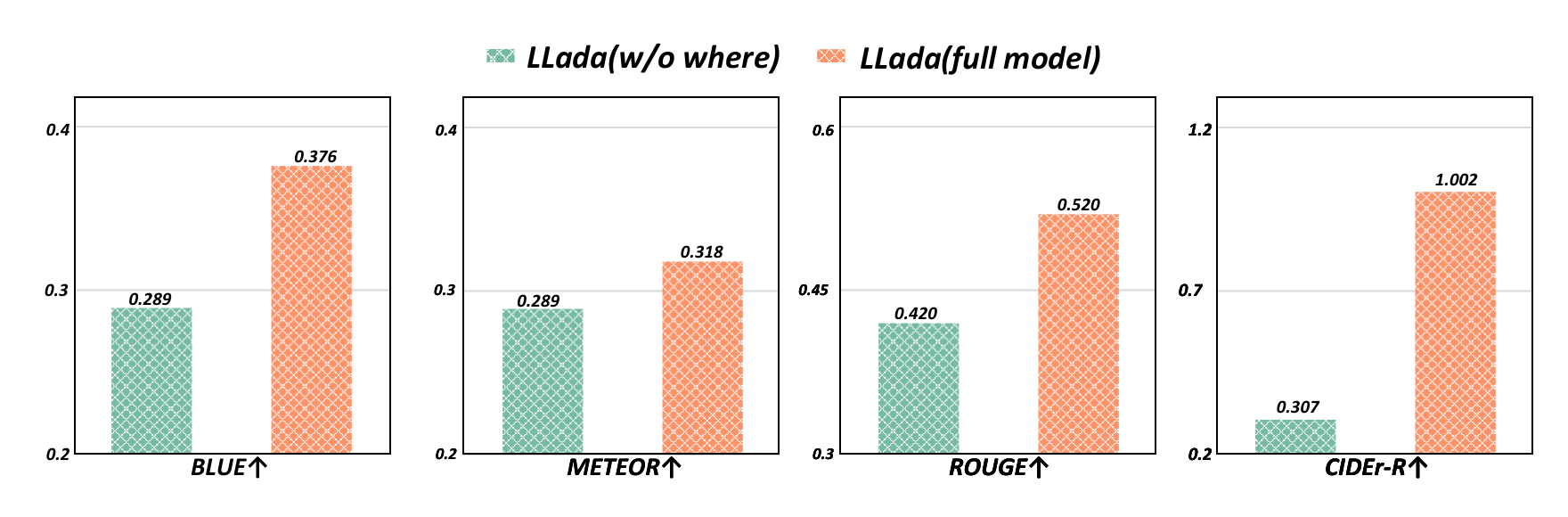}
        \caption{Impact of where on textual explanation generation.}
	\label{fig_abla_1}
\end{figure}

\begin{figure}[t]
        \centering
	\includegraphics[trim=5 5 5 5, clip=true, width=0.42\textwidth]{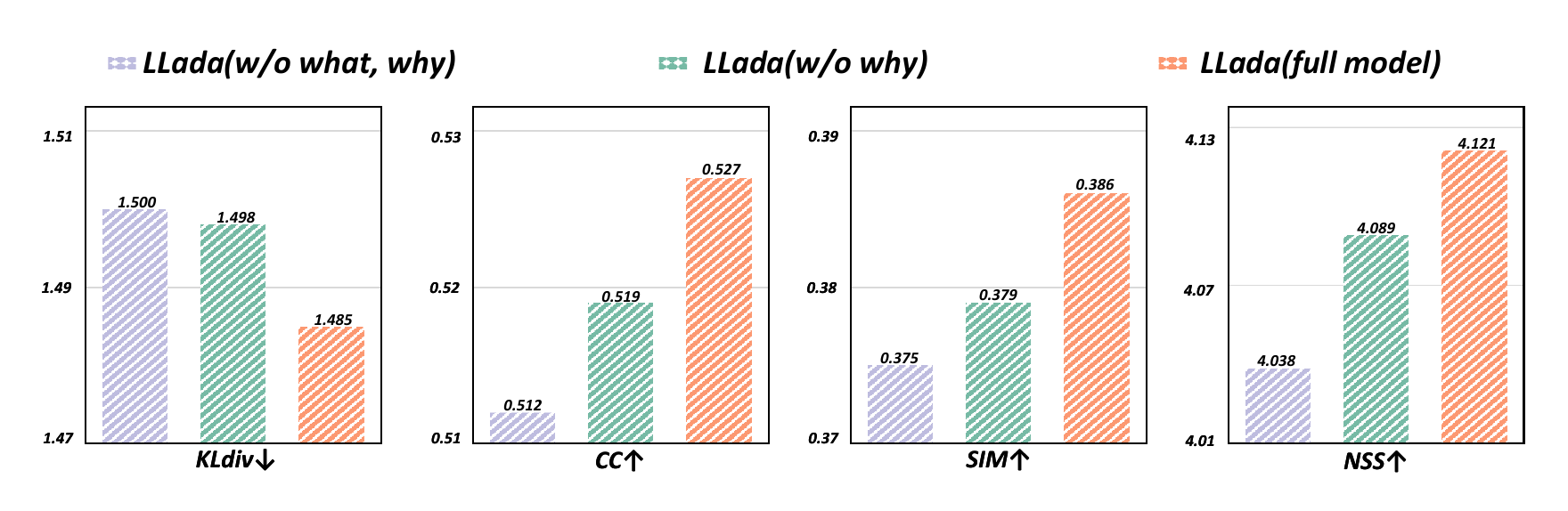}
        \caption{Impact of what and why on attention map prediction.}
	\label{fig_abla_2}
\end{figure}

To validate the effectiveness of LLada in jointly reasoning over \textit{where}, \textit{what}, and \textit{why}, we conduct ablation studies. First, we remove the \textit{where} task (attention map prediction) to assess its impact on \textit{what} and \textit{why} (textual explanation generation). As shown in Fig. \ref{fig_abla_1}, removing \textit{where} causes a significant performance drop in all textual explanation metrics. This demonstrates that \textit{where} plays a crucial role in \textit{what} and \textit{why} reasoning by providing essential spatial-level cues for more accurate and coherent explanations.

Next, we construct a baseline without \textit{what} and \textit{why} reasoning. We then progressively incorporate these tasks to assess their impact on the attention map prediction, as shown in Fig. \ref{fig_abla_2}. 
Introducing the \textit{what} task improves \textit{where} performance, indicating that textual descriptions provide valuable semantic cues for attention localization. Adding the \textit{why} task further boosts \textit{where} performance, confirming that causal reasoning strengthens attention map prediction.
Overall, these findings highlight the mutual reinforcement between \textit{where} and \textit{what/why} reasoning, particularly the strong facilitative effect of attention map prediction on textual explanation generation, demonstrating the effectiveness of LLada's joint optimization framework.

\iffalse
\begin{table}[t]
    \centering
    \small
    \setlength{\tabcolsep}{4pt}
    \caption{Impact of where on textual explanation generation.}
    \label{tab_ab_1}
    \resizebox{0.75\linewidth}{!}{ % 缩小表格
    \begin{tabular}{lcccc}
        \toprule
        Model & Bleu\_4↑ & Meteor↑ & Rouge↑ & CiderR↑ \\
        \midrule
        LLada (w/o where) & 0.289 & 0.289 & 0.420 & 0.307 \\
        LLada (full model) & \textbf{0.376} & \textbf{0.318} & \textbf{0.520} & \textbf{1.002} \\
        \bottomrule
    \end{tabular}
    }
\end{table}

\begin{table}[h]
    \centering
    \small
    \setlength{\tabcolsep}{4pt}
    %\caption{Impact of contextual information, what, and why on attention map prediction.}
    \caption{Impact of what and why on attention map prediction.}
    \label{tab_ab_2}
    \resizebox{0.65\linewidth}{!}{ % 缩小表格
    \begin{tabular}{lccc}
        \toprule
        Model & KLdiv↓ & CC↑ & SIM↑ \\
        \midrule
        LLada (w/o what, why) & 1.500 & 0.512 & 0.375 \\
        LLada (w/o why) & 1.498 & 0.519 & 0.379 \\
        LLada (full model) & \textbf{1.485} & \textbf{0.527} & \textbf{0.386} \\
        \bottomrule
    \end{tabular}
    }
\end{table}
\fi

%% file: sec/6_conclusion.tex
\section{Conclusion}
%Our work addresses the gap in existing driver attention models that fail to capture the cognitive motivations behind attention allocation in specific driving contexts, limiting in-depth understanding with explainability. To tackle this, we introduce Explainable Driver Attention Prediction, a novel paradigm that jointly predicts spatial attention regions (\textit{where}), parses attended semantics (\textit{what}), and provides cognitive reasoning for attention allocation (\textit{why}).
%We present W³DA, the first large-scale, explainable driver attention dataset, which enriches existing benchmarks with detailed semantic and causal annotations across diverse driving scenarios. Additionally, we propose LLada, a Large Language model-driven framework for driver attention prediction, which unifies pixel modeling, semantic parsing, and cognitive reasoning into an end-to-end architecture.
%Extensive experiments validate LLada’s effectiveness, demonstrating strong generalization across driving conditions. Our work pioneers a deeper understanding of task-driven attention mechanisms in driving, offering unprecedented insights for cognitive science, transparent autonomous systems, and human-computer interaction.

We introduce Explainable Driver Attention Prediction, a novel paradigm that extends beyond traditional pixel-wise attention modeling by jointly predicting \textit{where} drivers look, \textit{what} they attend to, and \textit{why} their attention is allocated. This enables a cognitive-level understanding of driver attention, bridging perception and reasoning in autonomous driving.
To support this paradigm, we present W³DA, the first large-scale explainable driver attention dataset, which significantly enriches existing benchmarks with detailed semantic and causal annotations across diverse real-world driving scenarios. Furthermore, we propose LLada, a Large Language Model-driven framework that unifies attention prediction, semantic parsing, and cognitive reasoning in an end-to-end architecture, offering human-like interpretability beyond conventional models. Extensive quantitative and qualitative analyses demonstrate LLada’s strong generalization across diverse driving conditions and datasets, highlighting the impact of cognitive cues in driver attention modeling. 
By systematically decoding the latent cognitive factors underlying attention allocation, our work pushes the frontier of task-driven attention modeling in driving, providing new insights for cognitive science, transparent autonomous systems, and human-computer interaction.

%% file: sec/7S_dataset.tex
\section{More Details about W³DA Dataset}
\label{sec:dataset}

% 数据标注pipeline （关键帧伪代码，数据集构建的prompt模板等）
% 关键帧的3个参数

\subsection{Dataset Comparisons}
\label{sec:dataset_com}

To highlight the advantages of W³DA, we compare it with existing driving attention datasets, as shown in Table~\ref{tab_dataset_comparison}. Unlike previous datasets that focus on single-domain scenarios with limited annotations, W³DA is the first to unify multiple driver attention benchmarks and introduce comprehensive multi-level annotations, enabling spatial (\textit{where}), semantic (\textit{what}), and cognitive (\textit{why}) reasoning in driver attention modeling. This framework provides a deeper understanding of driver attention mechanisms, offering new insights to the research community.

Existing datasets have significant limitations in diversity and annotation richness. DR(eye)VE \cite{Palazzi2019dreyeve} and LBW \cite{LBW} primarily focus on normal driving, while BDD-A \cite{xia2019predicting} and DADA-2000 \cite{fang2019dada, fang2022DADA}, despite covering safety-critical and accident scenarios, are collected in controlled environments. Moreover, previous datasets typically rely on uniform frame sampling (3 frames per second) to train models, which may introduce redundancy. Although most datasets provide pixel-level attention maps, none incorporate semantic or cognitive annotations, restricting their ability to explain the cognitive reasoning behind driver attention.

In contrast, W³DA not only spans normal, safety-critical, and accident scenarios, but also introduces an attention-aware key frame selection strategy, ensuring that extracted frames capture the most informative gaze patterns. Furthermore, W³DA is the first dataset to offer multi-level annotations across spatial, semantic, and cognitive dimensions, enabling interpretable attention modeling beyond simple heatmaps. By bridging the gap between spatial attention prediction and high-level reasoning, W³DA establishes a novel benchmark for explainable driver attention modeling.

\begin{table*}[h]
    \centering
    \caption{Comparison of W³DA with existing driving attention datasets.}
    \label{tab_dataset_comparison}
    \renewcommand{\arraystretch}{1.2}
    \resizebox{\textwidth}{!}{ % 自动调整表格大小
    \begin{tabular}{lccccccccc}
        \toprule
        \textbf{Dataset} & \textbf{Video Scenes} & \textbf{Frames} & \textbf{Sampling Method} & \textbf{Scene Domain} & \textbf{Collection} & \textbf{Public Access} & \textbf{Attention Map (Where)} & \textbf{Semantic (What)} & \textbf{Cognitive (Why)} \\
        \midrule
        DR(eye)VE~\cite{Palazzi2019dreyeve}  & 74   & 52,723   & Uniform Sampling & Normal Driving & On-road  & \cmark  & Pixel-level  & \xmark & \xmark \\
        BDD-A~\cite{xia2019predicting}  & 1,435 & 44,320   & Uniform Sampling & Safety-Critical & In-lab  & \cmark  & Pixel-level  & \xmark & \xmark \\
        DADA-2000~\cite{fang2019dada, fang2022DADA}  & 1,965 & 64,936   & Uniform Sampling & Accident & In-lab  & \cmark  & Pixel-level  & \xmark & \xmark \\
        LBW~\cite{LBW}  & 74   & 12,170   & Uniform Sampling & Normal Driving & On-road  & \cmark  & Pixel-level  & \xmark & \xmark \\
        ETOD~\cite{qin2022id}  & 16   & 6,487    & Uniform Sampling & Normal Driving & In-lab  & \xmark  & Instance-level  & \cmark & \xmark \\
        \midrule
        \textbf{W³DA (Ours)} & \textbf{3,548}  & \textbf{69,980}  & \textbf{Key Frames} & \textbf{Normal, Safety-Critical, Accident} & \textbf{On-road \& In-lab}  & \textbf{\cmark}  & \textbf{Pixel-level} & \textbf{\cmark} & \textbf{\cmark} \\
        \bottomrule
    \end{tabular}
    } % ← 这里补上
\end{table*}

\subsection{Dataset Statistics}
\label{sec:dataset_sta}

% 数据统计
% 数据集对比饼图->词云
In this subsection, we systematically analyze our W³DA dataset, including its composition in terms of data sources, scene categories, weather conditions, geographic locations, time periods, mean fixation maps, high-frequency semantic labels, and high-frequency cognitive reasoning causes. These diverse compositions ensure that the W³DA dataset provides a comprehensive representation of driving scenarios for attention prediction and analysis.

%\noindent \textbf{Data Sources \& Categories.}
\subsubsection{Data Sources \& Categories}
\begin{figure*}[h]
\centering
	\includegraphics[trim=0 0 0 0, clip=true, width=0.8\textwidth]{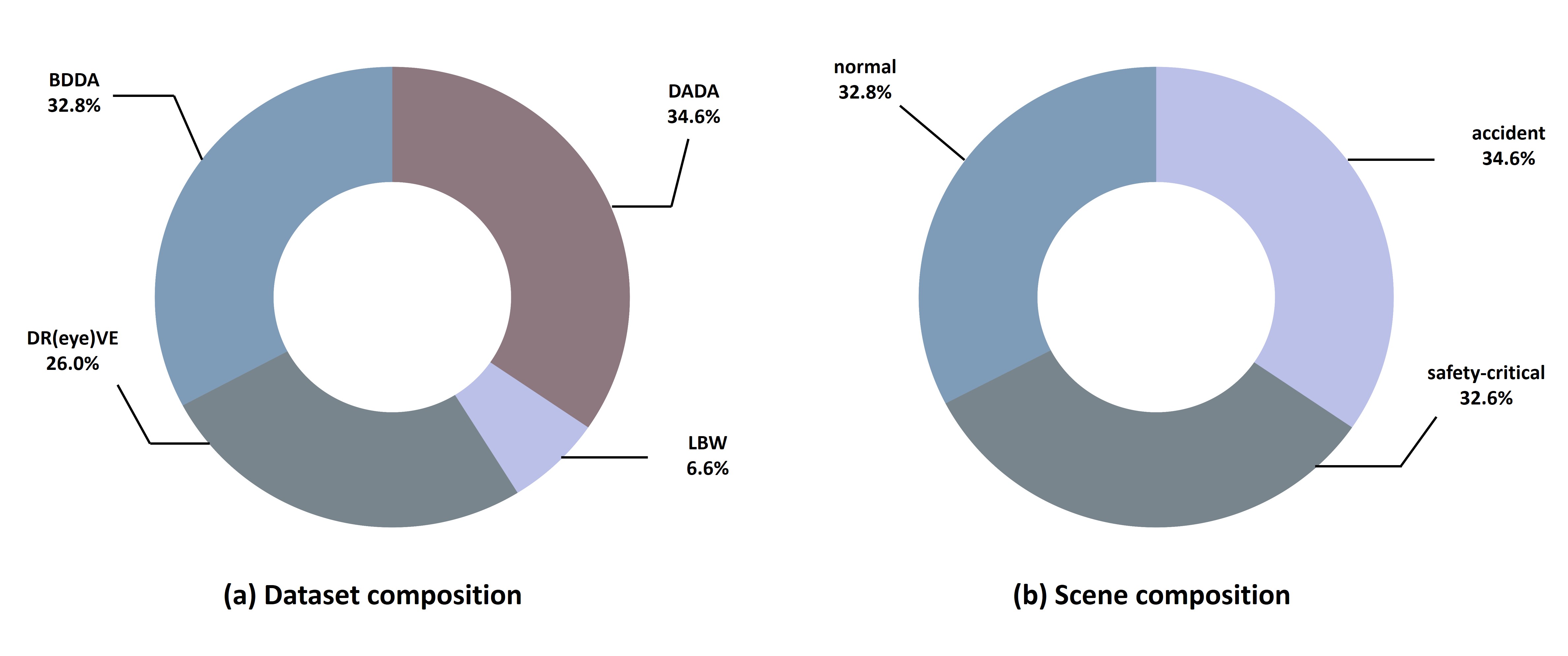}
	\caption{Data Sources \& Categories of W³DA.} 
	\label{fig_dataset1}
\end{figure*}

W³DA comprises 69,980 key samples extracted from 3,548 video scenes, covering diverse driving conditions. Specifically, it includes 22,839 normal driving samples from DR(eye)VE \cite{Palazzi2019dreyeve} and LBW \cite{LBW}, 22,950 critical situation samples from BDDA \cite{xia2019predicting}, and 24,191 traffic accident samples from DADA-2000 \cite{fang2019dada, fang2022DADA}. As shown in Fig. \ref{fig_dataset1}(a), the dataset is composed of 34.6\% of samples from DADA-2000, 32.8\% from BDDA, 26.0\% from DR(eye)VE, and 6.6\% from LBW. Additionally, Fig. \ref{fig_dataset1}(b) illustrates the scene composition, where 34.6\% of the scenes represent traffic accidents, 32.6\% are classified as safety-critical situations, and 32.8\% correspond to normal driving conditions. 
These diverse scenes, across a variety of weather conditions, times of day, and locations, with associated annotations, are the result of efforts from prior works \cite{fang2019dada, fang2022DADA, 2024_IV_SCOUT, 2024_IV_data}.

\subsubsection{Environmental Factors of Driving Scenarios}
\begin{figure*}[h]
\centering
	\includegraphics[trim=0 0 0 0, clip=true, width=1.0\textwidth]{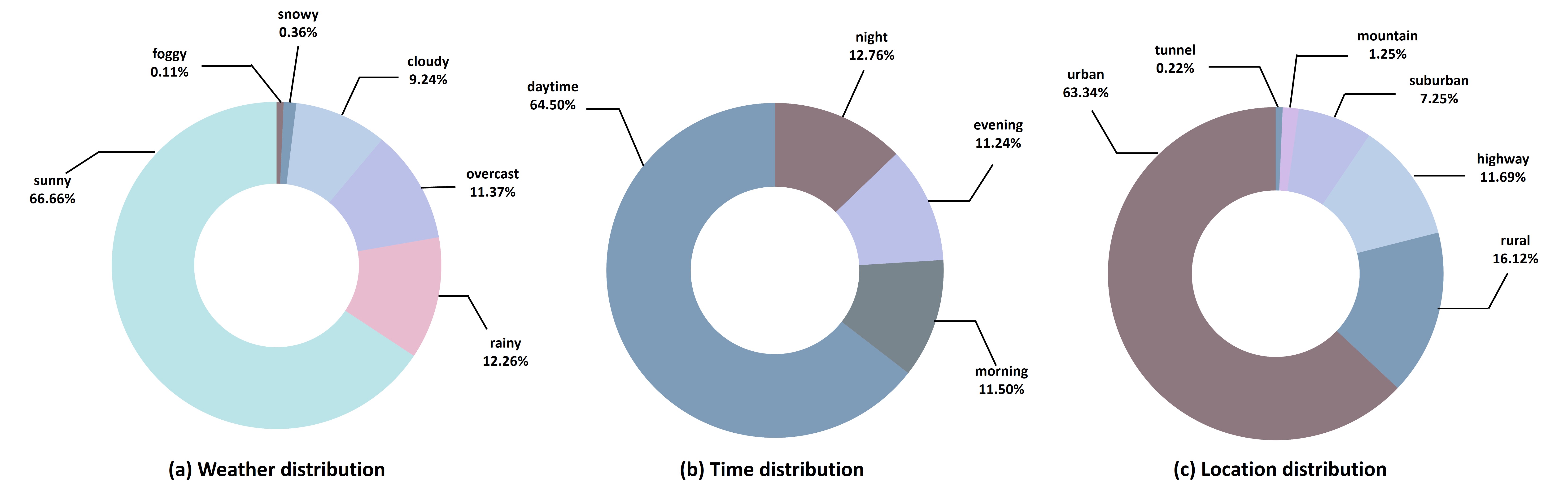}
	\caption{Distribution of weather conditions, time periods, and geographic locations in the W³DA dataset.} 
	\label{fig_dataset2}
\end{figure*}

\noindent \textbf{Weather Conditions.}
As shown in Figure~\ref{fig_dataset2}(a), the dataset includes a variety of weather conditions: sunny (66.66\%), rainy (12.26\%), overcast (11.37\%), and others such as snowy (0.36\%) and foggy (0.11\%). Sunny weather is the most common condition, representing the majority of the dataset. The inclusion of rainy, overcast, and extreme weather conditions (snowy and foggy) ensures that the dataset covers scenarios with different levels of visibility and driving challenges.

\noindent \textbf{Time Periods.}
Figure~\ref{fig_dataset2}(b) illustrates the dataset distribution across different times of day. Most samples are from daytime (64.50\%), followed by night (12.76\%), evening (11.24\%), and morning (11.50\%). % This distribution enables an analysis of how varying lighting and visibility conditions affect driver attention.

\noindent \textbf{Geographic Locations.}
The distribution of geographic locations is shown in Figure~\ref{fig_dataset2}(c). The dataset includes a diverse range of locations: urban (63.34\%), rural (16.12\%), highway (11.69\%), suburban (7.25\%), mountain (1.25\%), and tunnel (0.22\%). Urban areas are the most frequent location, which is typical for driving data collections due to the higher frequency of driving in these areas. However, the dataset also includes rural, highway, and less common environments like mountainous and tunnel regions, which offer additional driving challenges such as varying traffic density and road conditions.

\begin{figure*}[h]
\centering
	\includegraphics[trim=0 0 0 0, clip=true, width=0.75\textwidth]{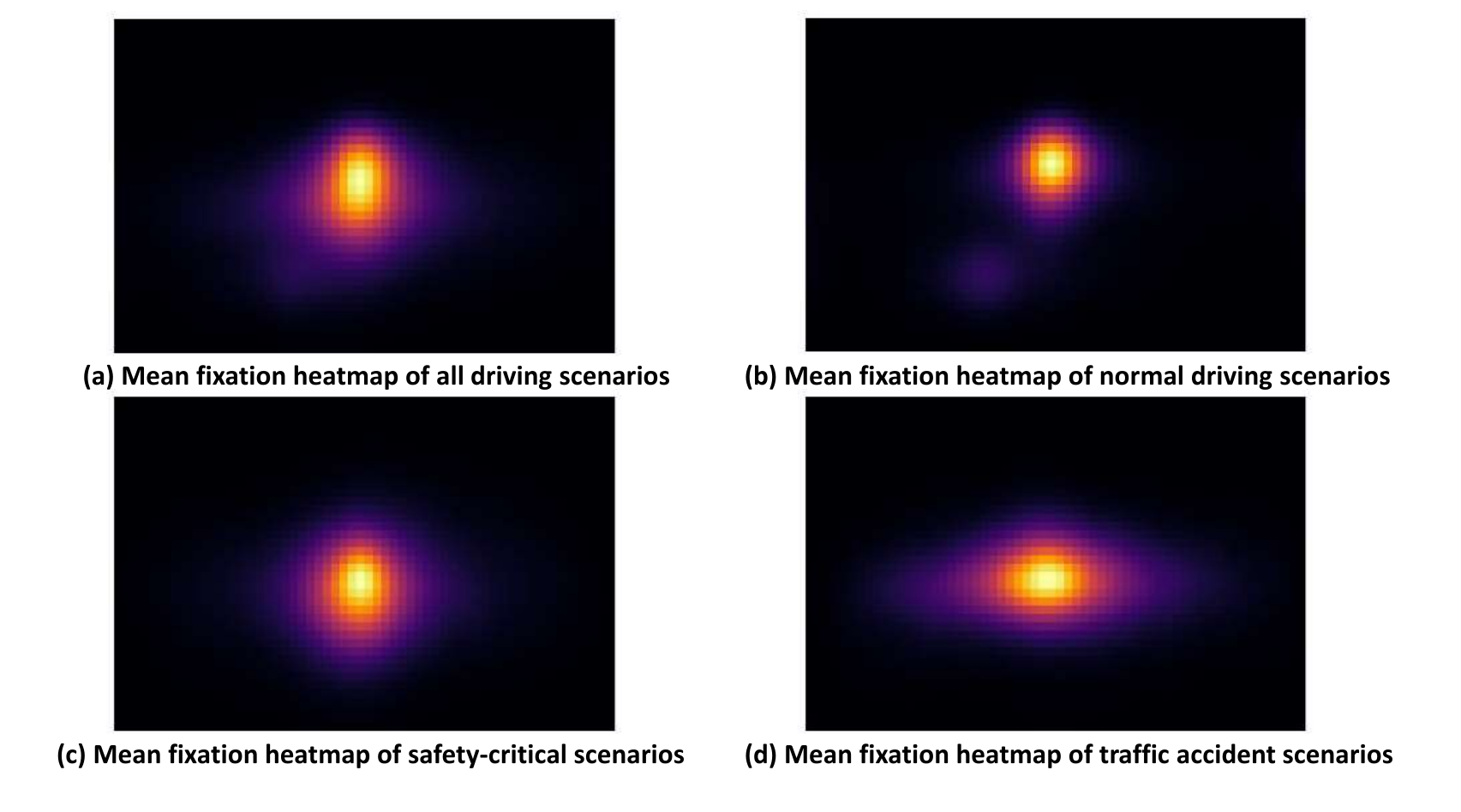}
    \caption{Mean fixation heatmaps for different driving scenarios.}
\label{fig_mean_heatmaps}
\end{figure*}

\subsubsection{Mean Fixation Maps (\textit{Where})}
To further analyze the overall distribution of driver attention in different driving scenarios, we compute the mean fixation maps across all samples in the dataset. Fig. \ref{fig_mean_heatmaps} presents the averaged fixation heatmaps for (a) all driving scenarios, (b) normal driving scenarios, (c) safety-critical scenarios, and (d) traffic accident scenarios.

\begin{figure*}[h]
\centering
	\includegraphics[trim=0 0 0 0, clip=true, width=0.7\textwidth]{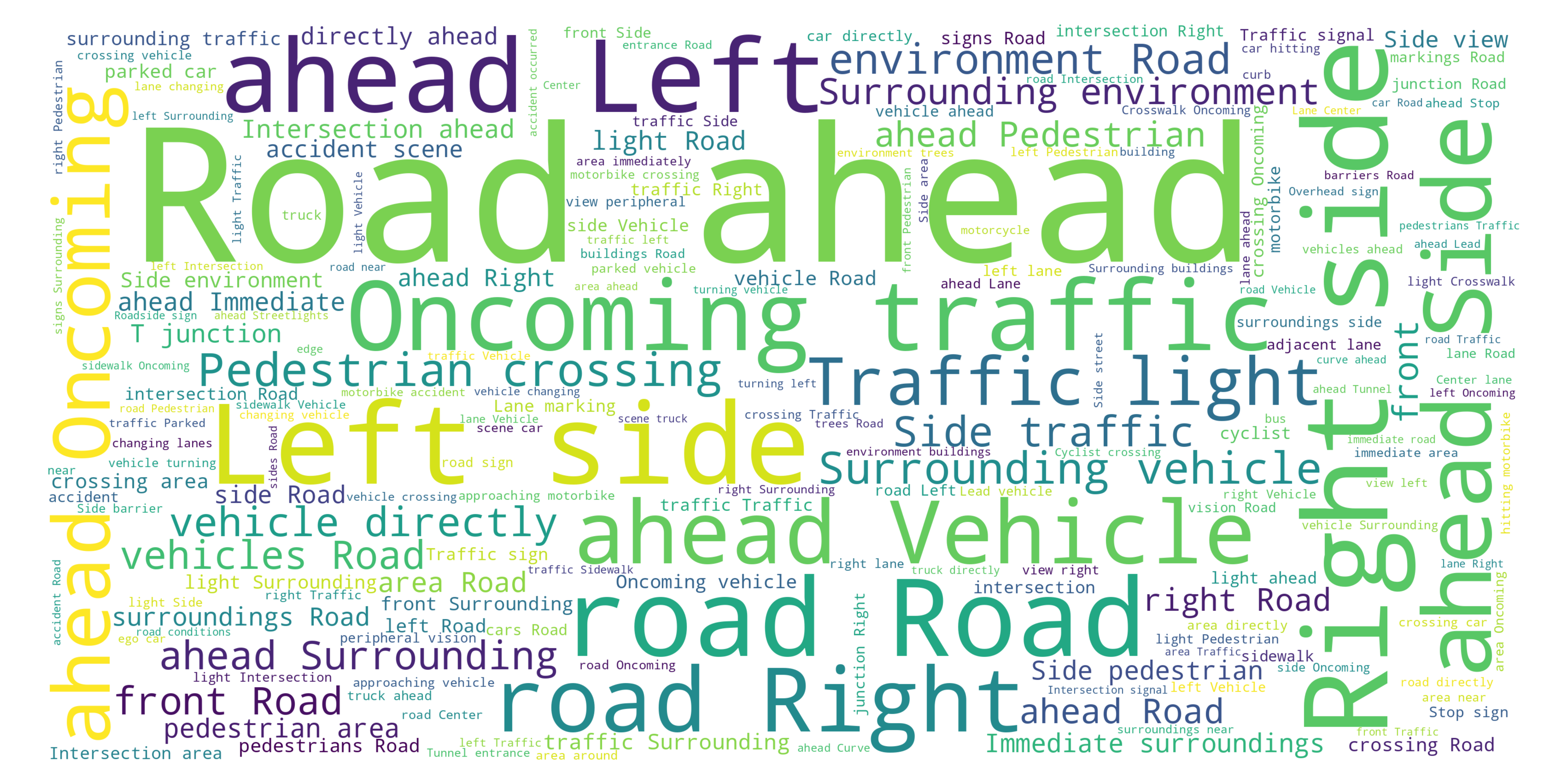}
	\caption{Word cloud of high-frequency semantic labels representing key elements in the driving scene, such as ``Road Ahead," ``Left side," ``Right side," ``Vehicle," ``Pedestrian, ``Traffic light," ``Oncoming traffic," and ``Intersection."}
	\label{fig_dataset3}
\end{figure*}

The fixation maps show that attention is predominantly allocated to the road ahead, aligning with the frequent occurrence of the label ``road ahead" in the semantic word cloud (Fig. \ref{fig_dataset3}). In normal driving, attention remains highly focused on the road center. In safety-critical scenarios, attention becomes more dispersed and shifts closer to the ego-vehicle, focusing on nearby vehicles and pedestrians that may impact driving safety. In traffic accident scenarios, fixation patterns exhibit a broader horizontal distribution, indicating active scene scanning to assess the accident situation and potential hazards.

%\noindent \textbf{High-frequency Semantic Labels (\textit{What}).}
\subsubsection{High-frequency Semantic Labels (\textit{What})}
As shown in the word cloud (Fig. \ref{fig_dataset3}), the most frequent labels include \textit{Road ahead, Left side of the road, Right side of the road, Vehicle, Pedestrian, Traffic light, Oncoming traffic, and Intersection}. These labels highlight critical regions that are essential for driver attention and decision-making.  

Among them, \textit{Road ahead} appears most frequently, reflecting the primary focus of drivers on the road directly in front of them. This is consistent with the fact that drivers typically prioritize the area ahead to assess the road conditions, potential obstacles, and upcoming traffic situations. The prominence of \textit{left side of the road} and \textit{right side of the road} suggests that lateral awareness, such as monitoring adjacent lanes and vehicles, is also crucial for safe driving. Other labels like \textit{Vehicle}, \textit{Pedestrian}, and \textit{Traffic light} further emphasize the importance of responding to dynamic elements and traffic signals.

\begin{figure*}[h]
\centering
	\includegraphics[trim=0 0 0 0, clip=true, width=0.7\textwidth]{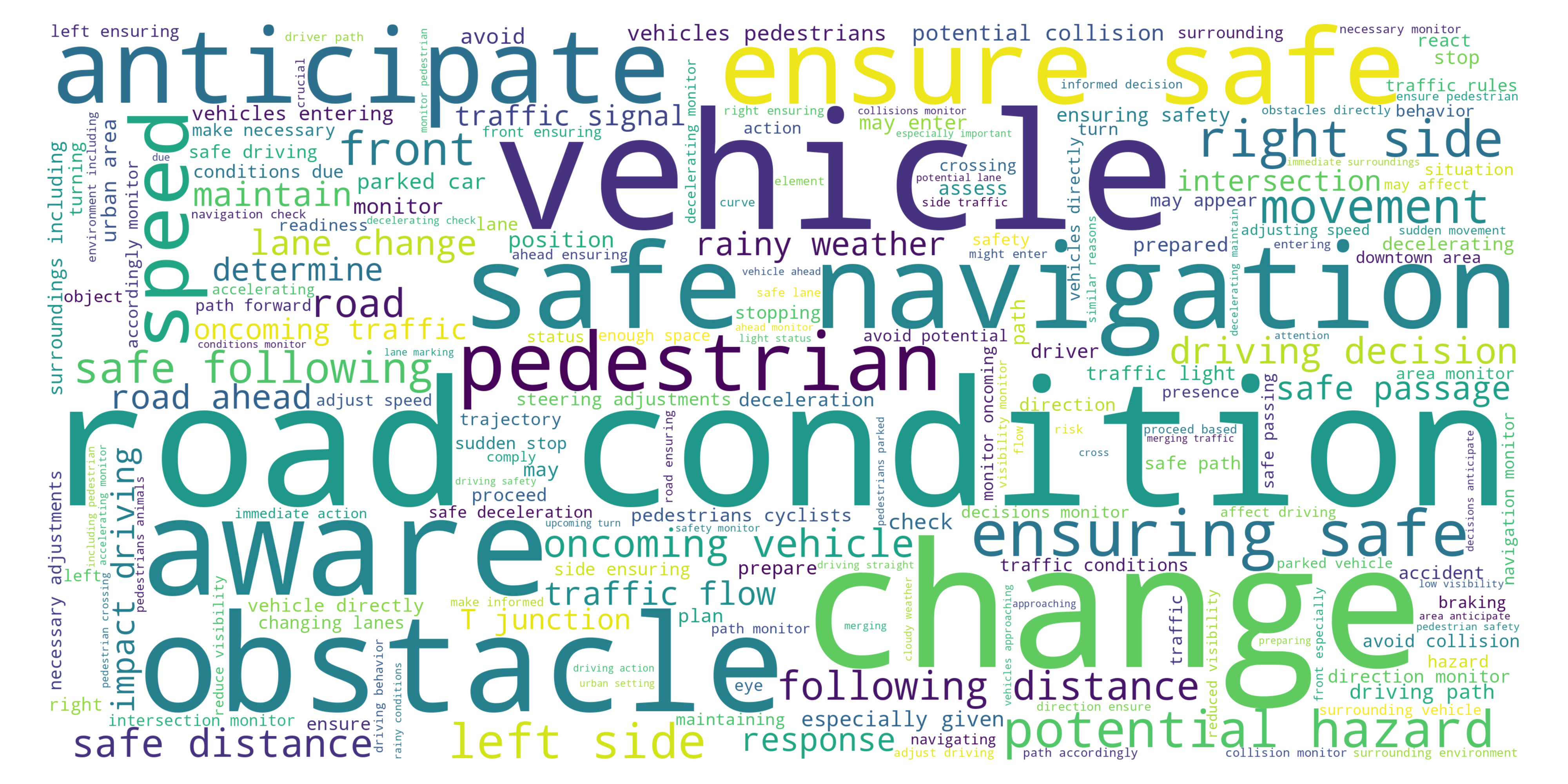}
    \caption{Word cloud of high-frequency cognitive causes, illustrating key factors such as ``road conditions," ``obstacle," ``safe navigation," ``safe distance," ``vehicle ahead," ``pedestrian," and ``change," which influence driver attention and decision-making in various driving scenarios.}
	\label{fig_dataset4}
\end{figure*}

%\noindent \textbf{High-frequency cognitive causes (\textit{Why}).} 
\subsubsection{High-frequency Cognitive Causes (\textit{Why})}
As shown in Fig. \ref{fig_dataset4}, the most frequent cognitive causes include labels such as \textit{road conditions, safe navigation, obstacle, change, safe distance, vehicle ahead}, and \textit{pedestrian}. These labels represent the primary factors that influence a driver's decision-making process and attention.

The high-frequency occurrence of labels such as \textit{road conditions, vehicle ahead, pedestrian}, and \textit{obstacle} reflects the driver’s safety awareness and perception of potential hazards. Labels like \textit{safe navigation} and \textit{safe distance} are driven by fundamental safety tasks, which ensure the driver maintains a secure path and appropriate spacing from other vehicles. The term \textit{change} highlights the significant impact of scene context variations, which influence attention allocation and driving decisions as the environment evolves with shifting traffic conditions or new obstacles. These cognitive causes play a crucial role in guiding the driver's actions, from adjusting speed to reacting to oncoming traffic and potential hazards, ensuring safety and informed decision-making on the road.

\subsection{More Details on Data Annotation}
\label{sec:dataset_anno}

\subsubsection{Key Sample Selection Details}

To complement our key sample selection strategy described in the main paper, we provide additional details on its implementation and parameterization.
Our selection process is based on three key metrics:  
(1) semantic similarity of driving scenes,  
(2) spatial divergence of driver attention, and  
(3) semantic similarity of driver attention.  

\noindent \textbf{Semantic Similarity of Driving Scenes:}
  The perceptual similarity between two consecutive frames is measured using the cosine similarity of their global scene embeddings extracted from the CLS token of the CLIP-Large image encoder \cite{radford2021learning}:
  \begin{equation}
  S_{\text{scene}}(I_t, I_{t-1}) = \frac{E_{\text{scene}}(I_t) \cdot E_{\text{scene}}(I_{t-1})}{\|E_{\text{scene}}(I_t)\| \|E_{\text{scene}}(I_{t-1})\|},
  \end{equation}
  where \( E_{\text{scene}}(I) \) represents the CLIP-extracted feature vector of image \( I \). A lower similarity score indicates a significant scene change (e.g., entering an intersection, encountering pedestrians).

% SPADA (Spatial Divergence of Driver Attention):  
\noindent \textbf{Spatial Divergence of Driver Attention:}
  The shift in spatial attention distribution is quantified using KL divergence between attention heatmaps:
  \begin{equation}
  D_{\text{KL}}(A_t \| A_{t-1}) = \sum A_t \log \frac{A_t}{A_{t-1}},
  \end{equation}
  where \( A_t \) represents the normalized attention map at time \( t \). A higher KL divergence indicates a significant change in spatial attention, often triggered by dynamic elements such as pedestrians or sudden braking events.

% SSDA (Semantic Similarity of Driver Attention):
\noindent \textbf{Semantic Similarity of Driver Attention.}
  To measure the semantic shift in attended regions, we compute the cosine similarity between attended region embeddings in consecutive frames:
  \begin{equation}
  S_{\text{attn}}(I_t, I_{t-1}) = \frac{E_{\text{attn}}(I_t) \cdot E_{\text{attn}}(I_{t-1})}{\|E_{\text{attn}}(I_t)\| \|E_{\text{attn}}(I_{t-1})\|},
  \end{equation}
  where \( E_{\text{attn}}(I) \) is the CLIP embedding of the attended region. A lower similarity score indicates a contextual shift in attention (e.g., transitioning from monitoring a traffic signal to checking pedestrians). 
  
  In practice. threshold values for each metric were tuned to balance redundancy reduction and keyframe retention: \( S_{\text{scene}} \leq 0.9 \), \( D_{\text{KL}} \geq 5 \), and \( S_{\text{attn}} \leq 0.9 \).

% \noindent \textbf{Threshold Selection.}

\subsubsection{Prompt Template for MLLM Annotation}

\begin{figure*}[h]
\centering
    \includegraphics[trim=0 5 0 5, clip=true, width=0.9\textwidth]{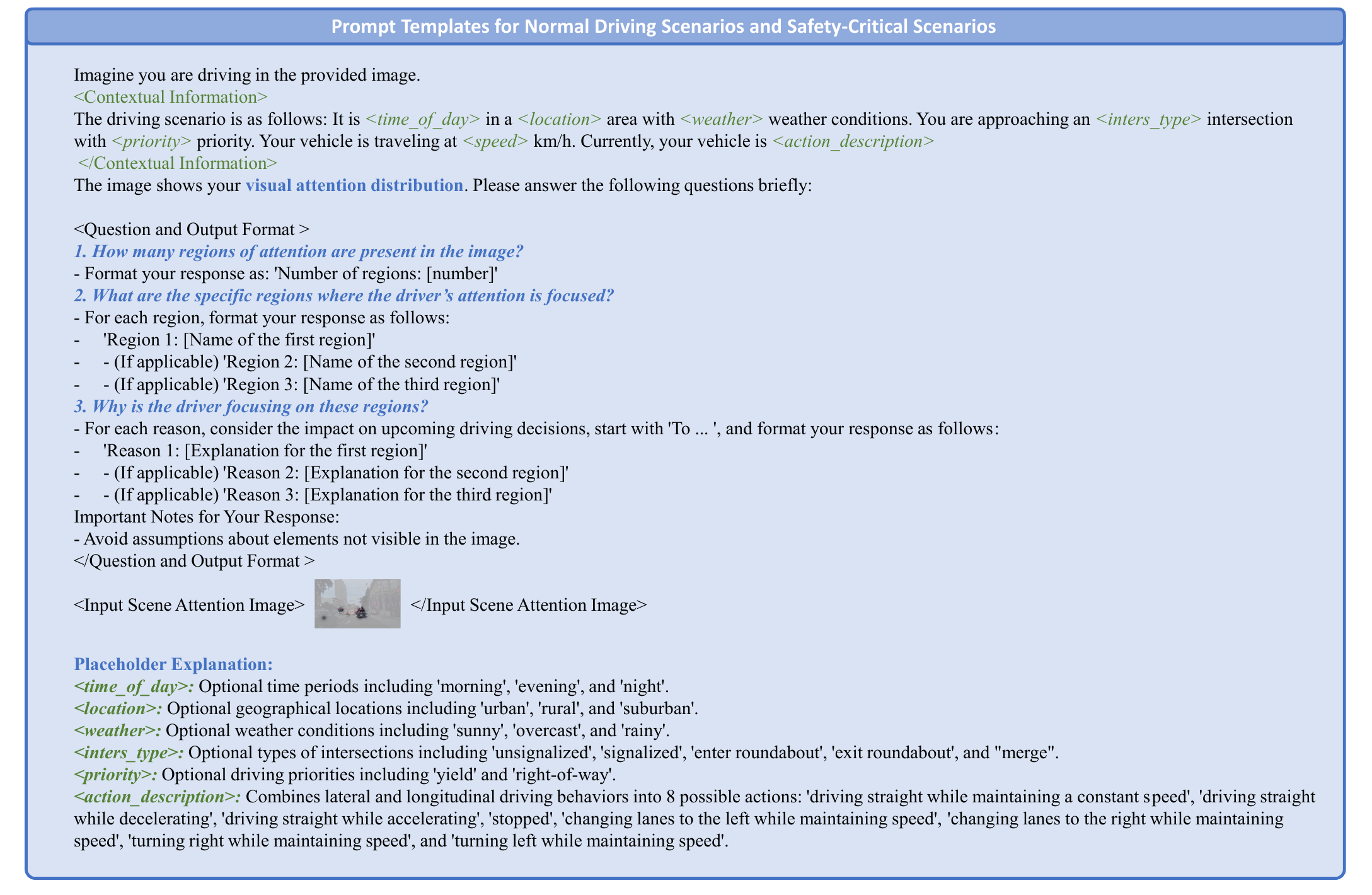}
    \caption{Prompt templates for MLLM annotation in normal and safety-critical driving scenarios.}
    \label{fig_prompt1}
\end{figure*}

\begin{figure*}[h]
\centering
    \includegraphics[trim=0 5 0 5, clip=true, width=0.9\textwidth]{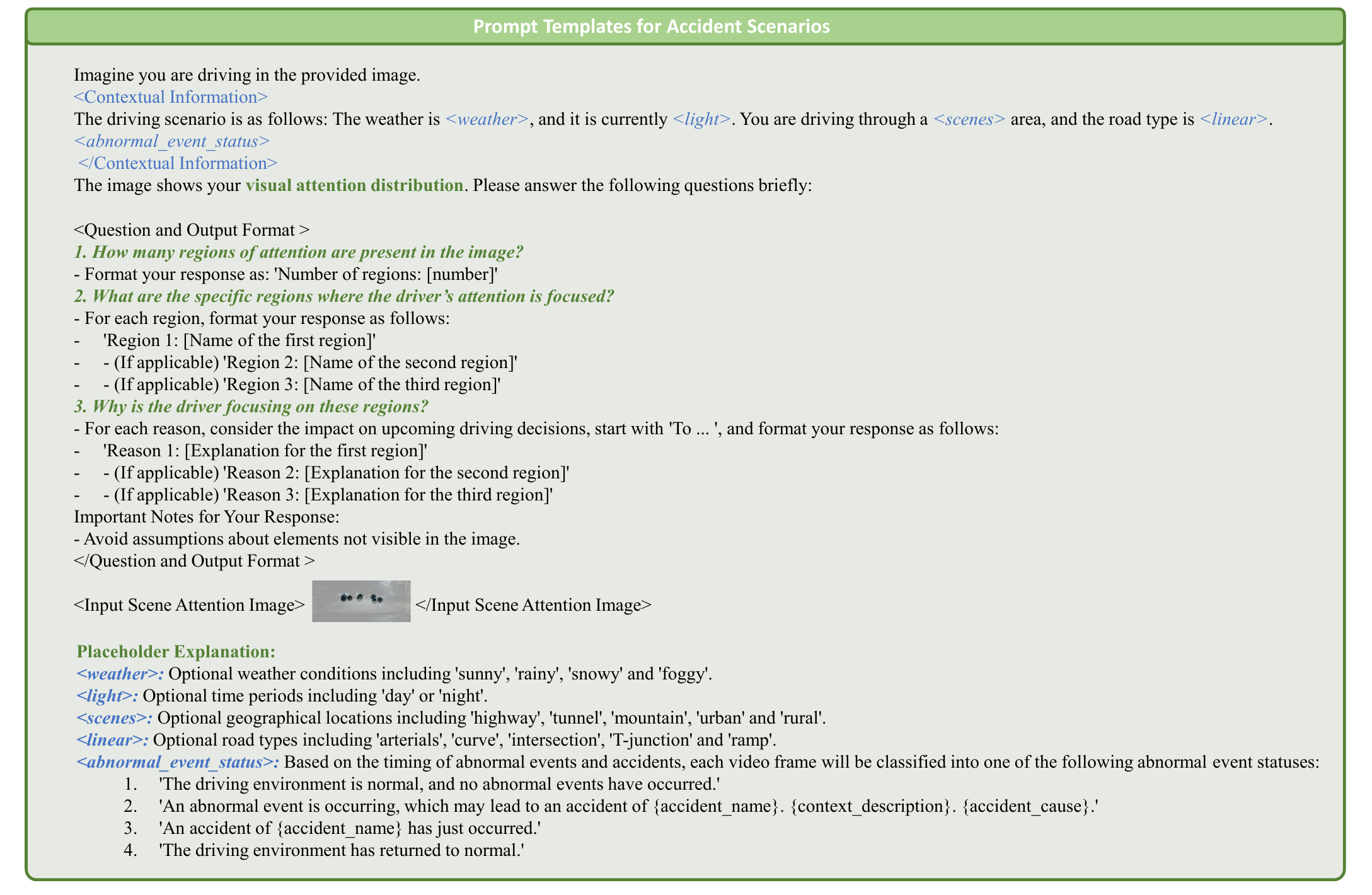}
    \caption{Prompt templates for MLLM annotation in traffic accident scenarios.}
    \label{fig_prompt2}
\end{figure*}

Here, we provide the prompt template used in our dataset annotation pipeline. Fig. \ref{fig_prompt1} illustrates the prompt template for normal and safety-critical driving scenarios, while Fig. \ref{fig_prompt2} presents the template for accident scenarios. The template guides the MLLM through a structured sequence of three tasks: identifying the number of attended regions, describing these regions, and explaining the reasons for attention. Additionally, it enforces a predefined response format, ensuring consistent and structured outputs for both the ``what" and ``why" aspects. 

\noindent \textbf{Contextual Prompts.}
The template incorporates rich contextual parameters, including weather conditions, time periods, geographic location, driving speed, right-of-way status, and road type, providing essential reference information for the MLLM. These parameters are embedded as placeholders within the template and are dynamically populated with scenario-specific data from each video frame, enabling frame-by-frame prompt construction in W³DA.

\noindent \textbf{Visual Prompts.}
For visual prompts, extensive preliminary experiments confirm the effectiveness of grayscale mask attention maps. Compared to heatmap-based visualizations, grayscale masks more effectively highlight critical visual information while preserving essential background details necessary for MLLM-based region descriptions and explanations. This advantage is particularly evident in low-visibility conditions, such as nighttime or rainy environments, where heatmap representations tend to obscure key details, exacerbating hallucination issues in MLLM responses.

\subsubsection{Human Verification}

% 尽管用于标注的MLLM在视觉理解及推理方面具有卓越的表现，但面对复杂的可解释驾驶注意力的数据生成还是会潜在有出现幻觉、不够准确等可能。主要问题包括：驾驶关注区域的数量不匹配,驾驶关注区域的语义描述不够准确,以及对应的因果原因解释不贴合场景上下文。To address this, we involve human experts to review and refine the model’s outputs based on their domain expertise, ensuring the final annotations are accurate, reasonable, and contextually aligned with the driving scenario.我们要求人类专家按照以下原则进行数据检查和refine：（1）注意数量及语义匹配：对区域描述中的幻觉实体进行过滤，对不准确或存在歧义的描述进行修正，并添加场景中实际存在但被MLLM忽略的；（2）注意因果原因与场景上下文匹配：对有悖于场景上下文的解释内容进行修正,确保解释符合当前场景下的实际驾驶目标；（3）所生成的因果解释符合在给定驾驶上下文中人类驾驶员直觉以及交通规则知识。上述过滤过程有效地确保了W3DA数据集中的语义及因果标签的高质量。

Although the proprietary MLLM API used for annotation demonstrates strong capabilities in visual understanding and reasoning, generating complex explainable driver attention data still poses challenges, including potential hallucinations and inaccuracies. To ensure the high quality of our dataset annotations, we incorporate human experts to review and refine the model’s outputs based on their domain expertise. This verification process guarantees that the final annotations are accurate, reasonable, and contextually aligned with the driving scenarios. 

\begin{figure*}[h]
    \centering
    \includegraphics[trim=5 5 5 5, clip=true, width=0.85\textwidth]{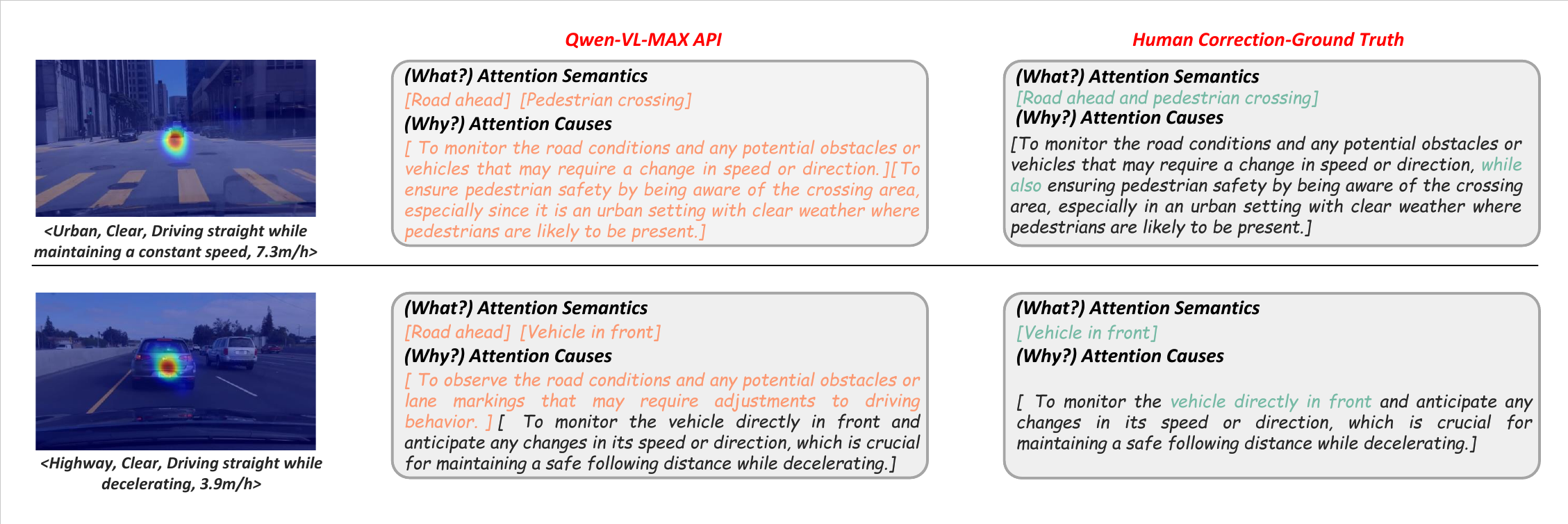}
    \caption{Comparison of MLLM-generated driver attention annotations (left) and human-corrected ground truth (right). The red text highlights hallucinated or inaccurate elements, while green text represents refined corrections ensuring contextual accuracy.}
    \label{fig_human_verification}
\end{figure*}

Figure~\ref{fig_human_verification} illustrates examples of MLLM-generated annotations compared to human-corrected ground truth. As shown, while the MLLM can accurately predict attended regions and reasoning explanations, it occasionally hallucinates non-existent objects or fails to capture critical contextual factors. Human verification ensures that these outputs are refined to better align with real-world driver cognition.

The process follows three key principles:

\begin{itemize}
 \item \textbf{Attention Validity and Completeness}:  
   Experts filter out hallucinated objects in attention region descriptions, refine inaccurate or ambiguous labels, and supplement missing yet contextually important entities that the MLLM failed to recognize. This ensures that the detected attention regions accurately reflect the \textit{true visual focus of human drivers} in each scenario.

\item \textbf{Contextual Causality and Driving Relevance}:  
   Experts correct explanations that contradict the \textit{scene context, driving objectives, and traffic dynamics}. This ensures that the \textit{``why" reasoning} remains \textit{logically coherent}, reflecting \textit{plausible cognitive processes of real drivers} rather than generic or overly rigid interpretations.

 \item \textbf{Human Intuition and Traffic Rule Adherence}:  
   The \textit{causal justifications} are further refined to align with \textit{real-world driving intuition and regulatory constraints}. This step ensures that the generated explanations are \textit{intuitively reasonable for human drivers} and \textit{comply with standard traffic laws}, avoiding unrealistic or noncompliant interpretations.
\end{itemize}

This verification and refinement process effectively guarantees the high quality of semantic and causal annotations in the W³DA dataset.

%% file: sec/8S_model.tex
\section{More Details about LLada model}
\label{sec_model}

% \subsection{Prompt Template}

% 在本章节，我们提供llada使用的prompt模板，如图10所示。LLada's input prompts share similarities with those used in dataset construction but differ in key aspects. First, LLada uses raw frame images as visual input rather than attention maps. Correspondingly, its text prompts exclude the phrase "The image shows your visual attention distribution." Furthermore, LLada's text prompts can accept arbitrary free-text format for contextual scene information rather than being constrained by placeholder templates.

\begin{figure*}[h]
\centering
    \includegraphics[trim=0 5 0 5, clip=true, width=0.9\textwidth]{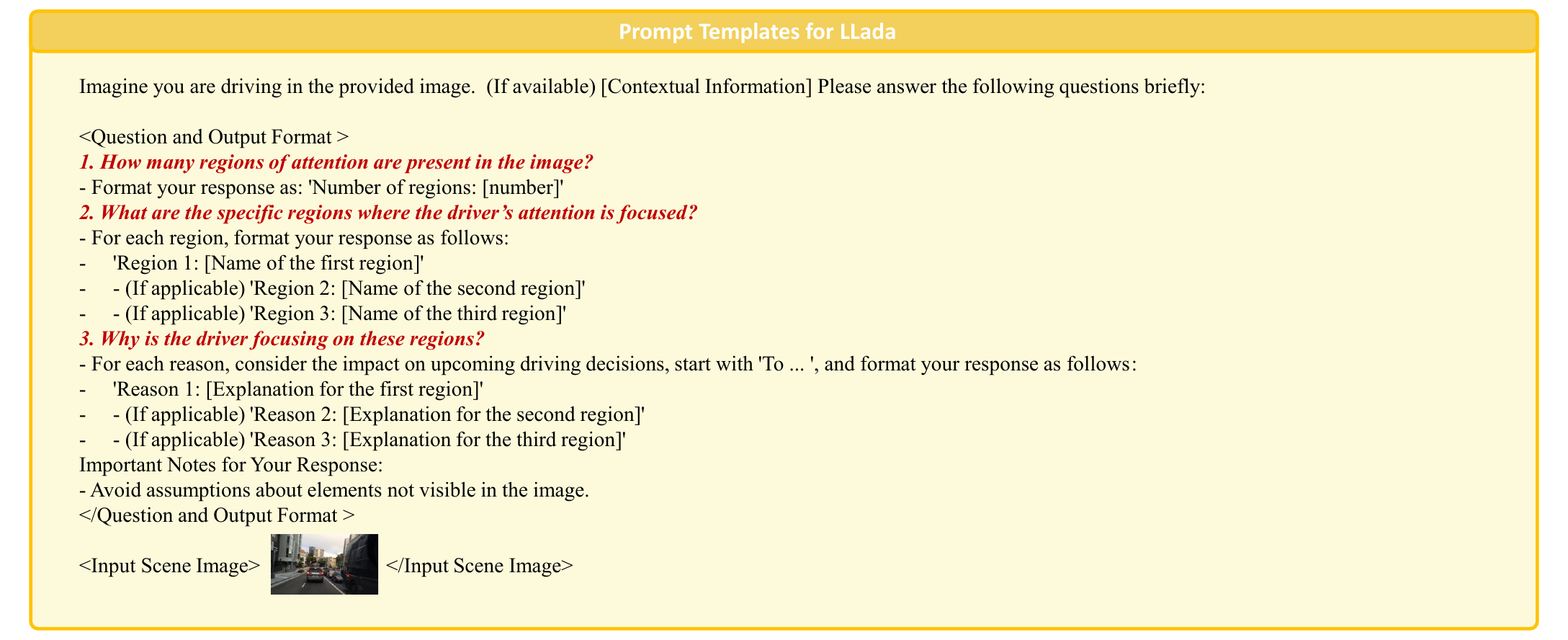}
    \caption{Prompt templates for LLada.}
    \label{fig_prompt3}
\end{figure*}

In this section, we provide details on the prompt templates used in LLada, as illustrated in Fig. \ref{fig_prompt3}. While LLada's input prompts share similarities with those used for dataset annotation, they differ in several key aspects.  
First, LLada takes raw frame images as visual input instead of attention maps. Consequently, its text prompts exclude the phrase ``The image shows your visual attention distribution." Additionally, unlike the structured placeholder-based templates used for dataset annotation, LLada allows flexible free-text input for contextual scene descriptions, enabling greater adaptability in diverse driving scenarios.

